\pdfoutput=1

\documentclass[11pt]{article}

\usepackage{pifont}  
\usepackage[table]{xcolor}
\newcommand{\cmark}{\textcolor{green!70!black}{\ding{51}}} 
\newcommand{\xmark}{\textcolor{red!80!black}{\ding{55}}}   

\usepackage[preprint]{acl}

\usepackage{times}
\usepackage{latexsym}

\usepackage[T1]{fontenc}
\usepackage{arydshln}

\usepackage[utf8]{inputenc}

\usepackage{microtype}

\usepackage{inconsolata}

\usepackage{booktabs}
\usepackage{multirow}
\usepackage{geometry}
\usepackage{makecell}

\usepackage{amsmath}
\usepackage{graphicx}
\usepackage{xparse}
\usepackage{amssymb}
\usepackage{float}

\usepackage{pifont}

\usepackage{array}
\usepackage{xcolor}
\usepackage{tcolorbox}

\title{No Universal Prompt: Unifying Reasoning through Adaptive Prompting for Temporal Table Reasoning}

\author{
    \textbf{Abhishek Rajgaria\textsuperscript{1}\thanks{These authors contributed equally.},
    \textbf{Kushagra Dixit\textsuperscript{1}\footnotemark[1]}\thanks{ Work done during internship at UPenn.}}, 
    \textbf{Mayank Vyas\textsuperscript{2}},
    \textbf{Harshavardhan Kalalbandi\textsuperscript{3}\footnotemark[2]}\\
    \textbf{Dan Roth}\textsuperscript{4},
    \textbf{Vivek Gupta\textsuperscript{2}\thanks{primary mentor corresponding author.}}\\
    \textsuperscript{1}University of Utah,
    \textsuperscript{2}Arizona State University,
    \textsuperscript{3} University of California Riverside,
    \textsuperscript{4}University of Pennsylvania\\
    \texttt{\small abhishek.rajgaria@utah.edu},
    \texttt{\small kushagra.dixit@utah.edu}, 
    \texttt{\small mvyas7@asu.edu}.
    \texttt{\small hkala002@ucr.edu}\\
    \texttt{\small danroth@seas.upenn.edu},
    \texttt{\small vgupt140@asu.edu}}

\begin{document}
\maketitle
\begin{abstract}
Temporal Table Reasoning is a critical challenge for Large Language Models (LLMs), requiring effective reasoning to extract relevant insights. Despite existence of multiple prompting methods, their impact on table reasoning remains largely unexplored. Furthermore, model performance varies drastically across different table and context structures, making it difficult to determine an optimal approach. This work investigates multiple prompting technique on diverse table types to determine that performance depends on factors such as \emph{entity type, table structure, requirement of additional context and question complexity}, with \emph{"NO"} single method consistently outperforming others. To address this, we introduce SEAR, an \textit{adaptive prompting} framework inspired by human reasoning that dynamically adjusts to context and integrates structured reasoning. Our results demonstrate that SEAR achieves superior performance across all table types compared to baseline prompting techniques. Additionally, we explore the impact of table structure refactoring, finding that a unified representation enhances model reasoning. 
\end{abstract}

\section{Introduction}

\begin{figure}[ht]
  \hspace{-0.2cm}
  \includegraphics[width=1.17\linewidth]{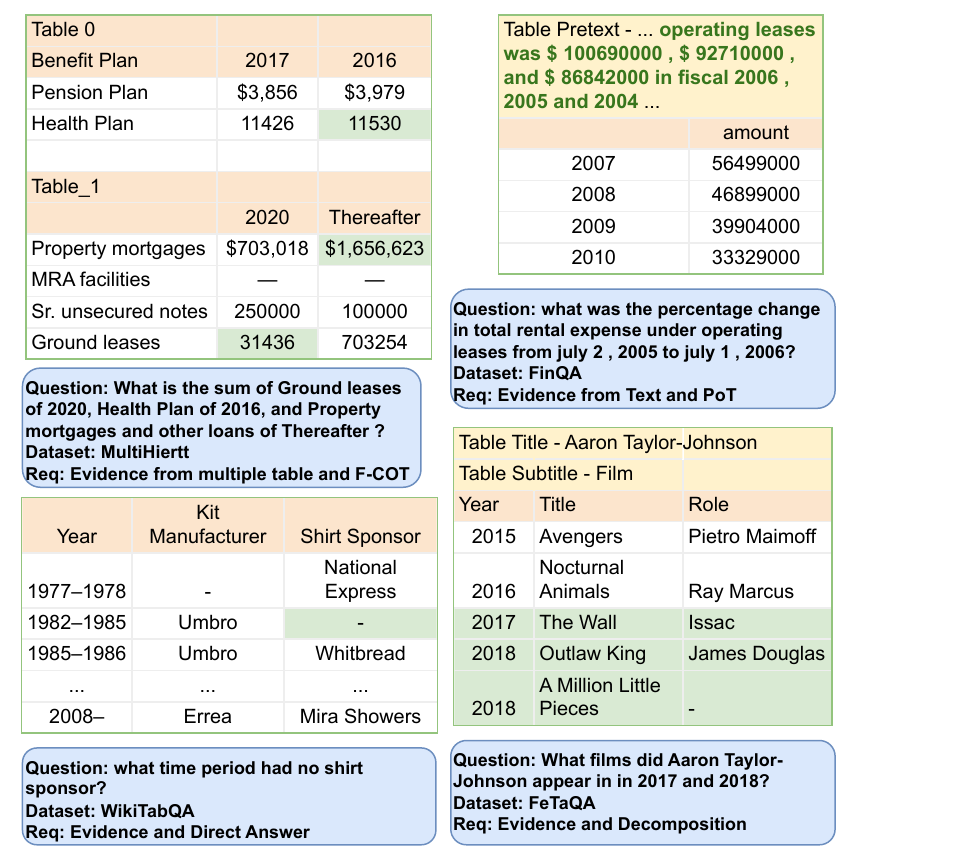}
  \caption{Examples of Different Table and Contextual structure, taken from different datasets with efficient reasoning method based on specific question, Full Tables are in Appendix \ref{sec: appendix_fig1_example}.}
  \label{fig:table_examples}
\end{figure}
Temporal table reasoning presents a unique challenge, requiring Large Language Models (LLMs) to interpret tabular data while capturing embedded temporal relationships. Unlike static tables that provide a fixed snapshot of information, temporal tables evolve over time, incorporating event sequences, timestamps, and dynamic updates. Reasoning over such structures is essential for tasks like financial forecasting, historical trend analysis, medical diagnosis, and event-based decision making \citep{gupta-etal-2023-temptabqa, xiong-etal-2024-large}. However, existing LLMs often struggle to model these intricate temporal dependencies, underscoring the need for more effective reasoning frameworks.


Recent work has demonstrated that LLMs can improve table reasoning performance through advanced prompting strategies \citep{10.1007/s11704-024-40330-z}. Nevertheless, studies such as \citet{wang-zhao-2024-tram} highlight persistent challenges in temporal reasoning, with models often failing to track evolving data or infer event sequences reliably. Moreover, most existing approaches rely on single-step prompting methods such as direct prompting or chain-of-thought reasoning \citep{10.5555/3600270.3602070} which frequently fail to generalize across diverse table structures and time-sensitive queries.

Although several prompting techniques have been proposed to improve LLM reasoning, their effectiveness for temporal table reasoning remains underexplored. In this study, we evaluate five single-step prompting methods as baselines Chain-of-Thought (CoT), Evidence Extraction, Decomposition, Faithful CoT \citep{radhakrishnan2023questiondecompositionimprovesfaithfulness}, and Program of Thought (PoT) \citep{chen2023programthoughtspromptingdisentangling}. Each baseline aims to enhance logical and numerical reasoning, yet their impact on temporal table reasoning performance has not been systematically analyzed. Furthermore, we evaluate an extensive set of baselines covering structural, temporal, and agentic reasoning approaches.

This study addresses this gap by analyzing the performance of multiple established baselines and a novel adaptive reasoning strategy on a Temporal Tabular Question Answering (TTQA) task.  
We aim to answer the following research questions: $(RQ1)$ Given a table and a question, which reasoning strategy should be employed?, $(RQ2)$ Is there a Single reasoning method that can perform well across all types of tabular structure?, and $(RQ3)$ Is there a unified representation that can encapsulate all different tabular structures in most effective manner for the TTQA task?

To address these research questions, we conducted experiments on eight distinct tabular structures using multiple state-of-the-art LLMs for the TTQA task. To overcome the limitations of existing baselines we propose \textbf{SEAR} (Select-Elaborate-Answer \& Reasoning) framework, a novel adaptive prompting strategy. Our motivation can be likened to a carpenter building a chair. They have many tools, such as a hammer, saw and drill. Each is capable of performing part of the task, but none of them can build the whole chair. It is the skillful selection and combination of these tools that brings the chair to life. Similarly, SEAR equips models with multiple reasoning tools, and then it is on the model's capability to choose them for solving the task at hand. From \ref{tab:reasoning_paths}, we observe that models actually use multiple tools to answer these questions.

SEAR operates in three distinct phases. In the Initial \textbf{Select} phase, it identifies high-level crucial steps, in the subsequent \textbf{Elaborate} phase it refines these steps by adding detailed instructions, ensuring comprehensive road map. Finally, the \textbf{Answer \& Reasoning} phase leverages the structured plan to deliver accurate answers, supported by clean, logical explanations and where necessary, includes integration of Python code for computational tasks.

Furthermore, we combined these three phases to create a single step reasoning strategy, which we call SEAR\_Unified. Our results show that SEAR\_Unified outperforms all single step baseline reasoning strategies by significant margins, and even standard 3-step SEAR and existing multi-step reasoning strategies such as Self Discover \cite{self_discover_zhou2024selfdiscoverlargelanguagemodels}. This demonstrates the supremacy and efficacy of our proposed reasoning strategy. Additionally, our study also includes detailed analysis of refactoring process, wherein we transform diverse tabular structure into a unified representation ("Refactor"),  Our main contributions are:

\begin{itemize}
\setlength\itemsep{0em}
    \item \textit{\textbf{Benchmarking Prompting Methods}}: We evaluate five single-step prompting methods and show that their effectiveness varies based on table structure, entity type, sparsity and question complexity.
    \item \textit{\textbf{Adaptive Reasoning Framework}}: We introduce SEAR, a multi-step adaptive prompting approach that generalizes well across diverse table structures, also we integrate them into a single unified adaptive prompt SEAR\_Unified, outperforming individual methods.
    \item \textit{\textbf{Table Structure Refactoring}}: We propose refactoring as an enhancement, demonstrating its effectiveness in improving model reasoning by optimized table representation.
    \item \textit {\textbf{Comprehensive Evaluation}}: We conduct a systematic analysis across various table types, highlighting the impact of different reasoning strategies and structure modifications.
\end{itemize}

\section{Why is Temporal Table Reasoning Challenging?}

Temporal table QA requires models to reason over structured data while accounting for time-dependent relationships. This challenge arises from three key factors: the diverse structures of tables, the domain-specific reasoning requirements, and the complexity of the questions asked.

\textbf{Structural Variability.}
Tables range from simple grids to hierarchical or semi‑structured layouts with merged cells and implicit links (e.g., HiTab’s multi‑level indexes, HybridQA’s tables mixed with text). They also come in diverse file formats (CSV, HTML, Markdown), so parsing must be flexible. SEAR first flattens and standardises these varied structures, making them easier for downstream reasoning.

\textbf{Domain-Specific Complexity.}
Reasoning strategies must adapt to the table's domain. Wikipedia-based datasets like WikiTableQuestions demand general factual reasoning and entity linking. Financial datasets like FinQA or TAT-QA emphasize numerical reasoning, requiring multi-step arithmetic and temporal trend analysis. SEAR dynamically adapts to these needs by identifying relevant entities and values, then applying suitable prompting strategies such as F-CoT or PoT. In numerically intensive domains, PoT facilitates executable code generation for precise computation.

\textbf{Question Complexity.}
Temporal QA questions range from direct lookups (e.g., “What year did the team win?”) to complex reasoning (e.g., “What was the profit two quarters after policy X?”). These often require temporal anchoring, arithmetic, and sequential logic. SEAR addresses this by decomposing questions and tailoring its strategy based on both table and query characteristics.

\begin{table*}[ht]
\centering
\scriptsize
\setlength{\tabcolsep}{4pt}
\begin{tabular}{l|ccc|cc|cc|ccc|cc}
\toprule
\textbf{Dataset} 
& \multicolumn{3}{c|}{\textbf{Structure}} 
& \multicolumn{2}{c|}{\textbf{Domain}} 
& \multicolumn{2}{c|}{\textbf{Reasoning}} 
& \multicolumn{3}{c|}{\textbf{Question Types}}
& \multicolumn{2}{c}{\textbf{Answer Types}} \\
& Flat & Hierarichal & Hybrid 
& Wikipedia & Finance. 
& Numerical & Textual 
& Lookup & Multi-step & Temporal & Long-form & SQL \\
\midrule
FeTaQA             & \cmark & \xmark & \xmark & \cmark & \xmark & \xmark & \cmark & \cmark & \cmark & \xmark & \cmark & \xmark \\
FinQA              & \cmark & \xmark & \xmark & \xmark & \cmark & \cmark & \cmark & \xmark & \cmark & \cmark & \xmark & \xmark \\
HiTab\textsuperscript{\textdagger}  & \xmark & \cmark & \xmark & \cmark & \cmark & \cmark & \xmark & \cmark & \cmark & \cmark & \xmark & \xmark \\
HybridQA           & \xmark & \xmark & \cmark & \cmark & \xmark & \cmark & \cmark & \cmark & \cmark & \xmark & \xmark & \xmark \\
MultiHierTT        & \xmark & \cmark & \xmark & \xmark & \cmark & \cmark & \cmark & \xmark & \cmark & \cmark & \xmark & \xmark \\
Squall             & \cmark & \xmark & \xmark & \cmark & \xmark & \cmark & \xmark & \cmark & \cmark & \xmark & \xmark & \cmark \\
TAT-QA             & \cmark & \xmark & \cmark & \xmark & \cmark & \cmark & \cmark & \cmark & \cmark & \cmark & \xmark & \xmark \\
WikiTableQuestions & \cmark & \xmark & \xmark & \cmark & \xmark & \cmark & \xmark & \cmark & \xmark & \xmark & \xmark & \xmark \\
\bottomrule
\end{tabular}
\caption{Comparison of Temporal Table QA datasets by structure, domain, reasoning, and question types. 
\textsuperscript{\textdagger}HiTab spans Wikipedia and financial domains. 
Binary indicators simplify complex question types (e.g., SQL, long-form).
}
\label{tab:datasets_matrix_final}
\end{table*}


\textbf{Limitations of Prior Work.}
Despite recent interest, most prior work underrepresents the structural and domain diversity seen in real-world tables. Datasets like TempTabQA~\cite{gupta-etal-2023-temptabqa} focus narrowly on specific formats, limiting generalizability. Annotation inconsistencies~\cite{deng2024enhancingtemporalunderstandingllms} further complicate benchmarking. Symbolic approaches (e.g., DATER~\cite{ye2023largelanguagemodelsversatile}, BINDER~\cite{cheng2023bindinglanguagemodelssymbolic}) offer logical precision on well-structured tables but falter on hybrid or semi-structured formats. Conversely, text-focused models (e.g., C.L.E.A.R.~\cite{deng2024enhancingtemporalunderstandingllms}) provide strong language understanding but lack robust symbolic reasoning. These limitations highlight the need for hybrid systems like SEAR, which dynamically integrate symbolic and neural strategies based on task demands.

\section{Adaptive Reasoning Framework}
\label{sec:adaptive_prompting}
Humans naturally  begin by understanding the objective and analyzing table structures, including cell relationships, headers, and implicit dependencies, while incorporating additional context if available. In temporal tables, this involves identifying both implicit and explicit time-based patterns. Once the problem and context are clear, relevant information is retrieved directly or by decomposing the task into subproblems based on complexity. Finally, logical and numerical reasoning is applied systematically to arrive at a well-founded conclusion.


Inspired by this intuitive approach, we propose the SEAR (Select-Elaborate-Answer \& Reasoning) a framework designed to dynamically adapt reasoning strategies based on the structure and complexity of the given table. SEAR builds upon existing prompting methods by introducing a structured, multi-step reasoning process that mirrors human problem solving. It follows a structured three step process to improve temporal table reasoning, ensuring systematic problem solving while leveraging In-context learning for adaptability.

    
\paragraph{Step1: Select Crucial Steps}: Identify key reasoning steps without answering directly, creating an efficient problem solving path. Figure \ref{fig:sears1_prompt} shows the actual prompt.
    \begin{itemize}
    \setlength\itemsep{0em}
        \item \underline{Problem Understanding}: Define the question's objective and analyze table structure.
        \item \underline{Reasoning Process}: Select single or multiple strategies from Extract relevant evidence, decompose complex queries, apply logical steps, and generate Python code if needed (when the question involves numerical or arithmetic reasoning. This is guided by the prompt as seen in Figure \ref{fig:sears1_prompt})
        \item \underline{Optimization tips}: Simplify steps, retrieve direct answers when possible, and use code for numerical operations.
    \end{itemize}

\paragraph{Step 2: Elaborate Crucial Steps}: Refine and comprehend selected steps for clarity and effectiveness. Figure \ref{fig:sears2_prompt} shows the actual prompt.
    \begin{itemize}
    \setlength\itemsep{0em}
        \item Add contextual details, specify exact table elements, and refine decomposition.
        \vspace{-5pt}
        \item Ensure a structured and logically coherent flow toward the final answer.
    \end{itemize}
    \paragraph{Step 3: Answer \& Reasoning}: Execute the structured steps to derive an accurate, well-supported answers. Figure \ref{fig:sears3_prompt} shows the actual prompt.
    \begin{itemize}
    \setlength\itemsep{0em}
        \item Follow elaborated steps precisely, referencing extracted evidence.
        \vspace{-5pt}
        \item Justify answers with logical explanations, when possible directly answer from evidence and integrate Python code for calculations when needed.
    \end{itemize}
    
By progressively refining reasoning, SEAR ensures adaptability and robustness across diverse table formats and complexities.


Standard SEAR is a three-step process that adds overhead and can impact efficiency. To address this, we propose SEAR\_Unified, a single-step adaptive prompt that merges SEAR’s structured reasoning into a unified framework. It dynamically selects and refines reasoning steps based on the query and table structure, retrieving key information, decomposing complex queries when needed, and selectively using Python for numerical operations. SEAR\_Unified validates intermediate steps and performs error checks to ensure accuracy while reducing redundant complexity. Figures \ref{fig:sear_prompt} and \ref{fig:sear_response} illustrate the prompt and reasoning path.


We also introduce table and context refactoring as a preprocessing step that clarifies headers, aligns data, and removes irrelevant context. This improves retrieval precision, reduces reasoning errors, and enhances adaptability across diverse tabular formats. Table \ref{tab:dataset_evaluation} summarizes the refactoring changes for each dataset.

\section{Experimental Setup}



\paragraph{Datasets.} We selected eight diverse tabular as shown in table \ref{tab:datasets_overview_short} datasets spanning structured, semi-structured, hierarchical, and hybrid tables to ensure a comprehensive evaluation. These datasets present challenges such as entity relations, numerical reasoning, and textual integration, making them well-suited for assessing table reasoning in LLMs as shown in Table \ref{tab:datasets_matrix_final}. For detailed overview of the dataset refer appendix \ref{dat: overview of dataset}.
\begin{table}[!htp]\centering
\scriptsize
\setlength{\tabcolsep}{0.5pt}
\begin{tabular}{lrrrrrrrr}
\toprule
\textbf{Categories} & \textbf{fetaqa} & \textbf{finqa} & \textbf{hitab} & \textbf{hybridqa} & \textbf{multi} & \textbf{squall} & \textbf{tatqa} & \textbf{wiki} \\\midrule
\textbf{Table Structure} & 1580 & 961 & 616 & 1528 & 1587 & 774 & 2240 & 1503 \\
\textbf{Title Clarity} & 1582 & 962 & 386 & 1528 & 1587 & 774 & 2244 & 1504 \\
\textbf{Column/Row Header} & 1268 & 919 & 353 & 1229 & 1587 & 774 & 2158 & 1283 \\
\textbf{Data Formatting} & 1329 & 957 & 269 & 1476 & 1585 & 774 & 2124 & 1399 \\
\textbf{Bolding \& Emphasis} & 1207 & 934 & 206 & 1460 & 1524 & 347 & 2200 & 478 \\
\textbf{Other} & 328 & 273 & 82 & 468 & 539 & 197 & 696 & 309 \\\bottomrule
\end{tabular}
\vspace{-0.75em}
\caption{\small Dataset evaluation for refactoring categories.}
\label{tab:dataset_evaluation}
\vspace{-1.5em}
\end{table}


\textit{Dataset Filtering:} Adapting TempTabQA’s \cite{gupta-etal-2023-temptabqa} keyword filter (§3.2), we selected temporal cues (e.g., before, year, latest) along with domain-specific terms (e.g., fiscal, quarterly) and applied them across all datasets. This approach reliably captures most of the explicit temporal questions, though purely implicit cases may be missed. Incorporating human judgment could improve coverage but at the cost of scalability.

\newcolumntype{R}{>{\raggedleft\arraybackslash}p{8mm}}   

\begin{table}[ht]
\centering
\scriptsize
\setlength{\tabcolsep}{3pt}  
\begin{tabular}{@{}l p{50mm} R@{}}
\toprule
\textbf{Dataset} & \textbf{Brief description} & \textbf{\#Qs} \\
\midrule
FeTaQA      & Wikipedia tables; long‑form answers from discontinuous facts & 1,582 \\
FinQA       & Financial reports; multi‑step numerical reasoning & 962 \\
HiTab       & Hierarchical tables; fine‑grained numeric questions & 897 \\
HybridQA    & Wiki tables + linked text; hybrid reasoning & 1,528 \\
MultiHierTT & Finance; multiple hierarchical tables + long text & 1,587 \\
Squall      & WikiTableQ + SQL alignments; structured query tasks & 774 \\
TAT‑QA      & Finance; tables + text with arithmetic / counting & 2,244 \\
WikiTableQ  & Wikipedia trivia; factual + numeric Q over large tables & 1,504 \\
\bottomrule
\end{tabular}
\caption{Number of retained temporal Questions.}
\label{tab:datasets_overview_short}
\end{table}
\vspace{-8pt}
\paragraph{Models:} We used 3 LLM models: GPT4o-mini, Gemini 1.5 Flash, and LLaMA 3.1 70B. 

\paragraph{Prompts \& Frameworks:} Effective prompting improves task comprehension and response quality by providing structured instructions. We evaluated 13 prompting strategies spanning direct, structured, temporal, and agentic approaches, as summarized in Table \ref{tab:baseline_categories}.

\begin{table*}[t]              
\centering
\scriptsize
\setlength{\tabcolsep}{4pt}
\newcolumntype{Q}{>{\raggedright\arraybackslash}p{75mm}} 
\begin{tabular}{@{}l Q l@{}}
\toprule
\textbf{Baseline} & \textbf{Brief description} & \textbf{Category} \\
\midrule
Chain‑of‑Thought (COT) \cite{10.5555/3600270.3602070}            & Step‑by‑step natural‑language rationale                                  & Direct \\
Evidence Extraction(EE)          & Extracts supporting cells first, then answers                            & Direct \\
Decomposed Prompting(Decomp)\cite{khot2023decomposedpromptingmodularapproach}     & Splits complex queries into simpler sub‑prompts                          & Direct \\
Faithful COT (F‑COT) \cite{lyu-etal-2023-faithful}  & Adds consistency checks to Chain‑of‑Thought                              & Direct \\
Program‑of‑Thought (POT)\cite{chen2023programthoughtspromptingdisentangling}           & Generates executable code (e.g., Python) for reasoning                   & Direct \\[2pt]

Self‑Discover   \cite{self_discover_zhou2024selfdiscoverlargelanguagemodels}                  & Model autonomously picks reasoning modules                               & Structured \\
Self‑Ask \cite{press-etal-2023-measuring_self_ask}                          & Iteratively asks and answers sub‑questions                               & Structured \\
Plan \& Solve  \cite{wang-etal-2023-plan}                   & Separates plan generation from execution                                 & Structured \\[2pt]

C.L.E.A.R. \cite{deng-etal-2025-enhancing}                        & Injects temporal cues for semi‑structured tables                         & Temporal \\
Narration of Thought (NoT) \cite{zhang-etal-2024-narrative}       & Requires chronological narration to keep temporal order                  & Tempooal \\[2pt]

Self‑Consistency Prompting (SCP) \cite{wang2023selfconsistency} & Samples multiple COTs and votes                                          & Agentic \\
Tree of Thought (ToT)  \cite{yao2023tree}           & Searches a tree of reasoning states with pruning                         & Agentic \\
Graph of Thought (GoT) \cite{besta2023graph}            & Generalises ToT to graph search                                          & Agentic \\
\bottomrule
\end{tabular}
\caption{Prompting baselines grouped by category.}
\label{tab:baseline_categories}
\end{table*}

To ensure a balanced evaluation, we included both textual and symbolic reasoning prompts. CoT, Evidence Extraction, and Decomposed Prompting guide models through step‑by‑step interpretation. SCP augments multiple chains of thought and selects the majority vote. PoT and F‑CoT generate structured logic for consistent reasoning. The temporal baselines NoT and C.L.E.A.R. inject explicit chronological cues to help the model track event ordering. The structured baselines Self‑Discover, Self‑Ask, and Plan \& Solve introduce autonomous decomposition and planning to improve reasoning quality. The agentic methods ToT and GoT explore tree‑ or graph‑structured reasoning paths to identify high‑value solutions. All methods were evaluated in a few‑shot setting except Self‑Discover.


\paragraph{Evaluation:} Evaluating diverse datasets is challenging due to varying answer types, from numerical values to long-form text. A rigid metric may miss semantic correctness, so we propose the \textit{\underline{Hybrid Correctness Score (HCS)}}, which balances lexical and semantic accuracy by combining Relaxed Exact Match Score (REMS, F1-based) and Contextual Answer Evaluation (CAE, LLM-based). A response is considered correct if its REMS score exceeds 80 or if CAE deems it correct. By integrating both lexical and contextual evaluation, HCS offers a more robust measure of answer correctness. \textbf{all reported scores represent HCS} for consistency. Detailed REMS and CAE results are provided in Tables \ref{tab:rems_cae_gpt} \ref{tab:rems_cae_gemini}, \ref{tab:rems_cae_llama} in Appendix \ref{sec:appendix_rems_cae}.

\section{Result and Analysis}

In this section, we analyze results using Tables (\ref{tab:hcs_gemini}, \ref{tab:hcs_gpt}, \ref{tab:hcs_llama}) which showcase HCS scores. 

\paragraph{Is there a single existing reasoning strategy which works best on all table types?} Performance varies depending on table structure, domain, and question complexity. As observed in Gemini 1.5 Flash results (Table \ref{tab:hcs_gemini}), COT performs best on HybridQA, Evidence Extraction excels in HiTab, TATQA, FeTaQA and Squall, while Decomposition is most effective for WikiTabQA and FinQA. POT shows the highest performance in MultiHierTT, whereas F-COT does not emerge as the best baseline in any dataset. A similar trend is evident across GPT and LLaMA models as shown in Table \ref{tab:best_baseline}. Thus, no single prompting method universally outperforms others, as effectiveness is higly dependent on the dataset's structure and complexity.

\begin{table}[!htp]\centering
\scriptsize
\begin{tabular}{lrrrr}\toprule
&Gemini 1.5 Flash &GPT 4o mini &Llama 3.1 70B \\\midrule
COT &HybridQA &MultiHierTT &HiTab \\
& &TATQA &HybridQA \\
& &FeTaQA & \\
\hline
EE &HiTab &WikiTabQA &FeTaQA \\
&TATQA &HiTab &Squall \\
&FeTaQA &HybridQA & \\
\hline
Decomp &Squall & & \\
&WikiTabQA &FinQA &WikiTabQA \\
&FinQA & &MultiHierTT \\
& & &TATQA \\
\hline
POT &MultiHierTT &Squall &FinQA \\
\hline
F-COT &- &- &- \\
\bottomrule
\end{tabular}
\caption{Dataset for which Baseline reasoning strategy performed best for each model}\label{tab:best_baseline}
\end{table}

\paragraph{Does the Adaptive Reasoning Framework Help?} Table \ref{tab:best_baseline} confirms that COT, Evidence Extraction, and Decomposition dominate in most datasets, with POT and F-COT experience improvement in performance for financial and Squall datasets. SEAR dynamically selects its reasoning path, primarily leveraging Evidence Extraction, Decomposition, and Logical Steps (COT) while integrating Python Program for numerical reasoning. by design, it optimally combines dominant reasoning strategies with computation support. SEAR outperforms baseline in 5 dataset for Gemini, in 2 dataset for GPT, and in 4 datasets for LLaMA. While SEAR consistently improves performance over baseline across multiple models, it does not generalize equally across all datasets. 

\paragraph{Does unification of SEAR help?}  SEAR\_Unified optimizes reasoning by merging and refining steps into a single adaptive prompt, reducing overhead while enhancing flexibility. As seen in Table \ref{tab:hcs_gemini}, \ref{tab:hcs_gpt} , \ref{tab:hcs_llama}, SEAR\_Unified outperforms baselines across all datasets for Gemini, while for GPT and LLaMA, it surpasses baselines in 6 datasets, demonstrating its superiority. This highlights SEAR\_Unified's ability to generalize effectively across diverse datasets and models.

We compared our methods with recent structured and modular reasoning approaches, including Self‑Discover, Self‑Ask, and Plan \& Solve. Our approach consistently outperforms these baselines, with particularly strong gains on Multi‑HierTT, HiTabs, Squall, and HybridQA. Among them, Self‑Discover performs the closest, underscoring the value of modular and adaptive reasoning. We also benchmarked against temporal (NoT, C.L.E.A.R.) and agentic (ToT, GoT, SCP) strategies. Although NoT, C.L.E.A.R., and GoT perform well on FetaQA, TAT‑QA, and HiTabs, they fail to deliver consistent improvements on more complex benchmarks.

\begin{table}[!htbp]\centering
\scriptsize
\setlength{\tabcolsep}{1.7pt}
\begin{tabular}{lccccccccc}\toprule
&wiki &multi &hitab &finqa &tatqa &fetaqa &squall &hybridqa \\\midrule
COT &73.60 &58.79 &79.04 &60.08 &87.30 &71.30 &69.90 &80.76 \\
F-COT &66.89 &60.68 &52.06 &62.16 &78.79 &56.13 &61.11 &17.93 \\
Decomp &78.52 &61.00 &75.47 &62.58 &91.67 &67.07 &67.57 &74.67 \\
EE &76.33 &60.43 &80.82 &55.93 &92.20 &77.62 &72.32 &80.10 \\
PoT &74.40 &61.12 &70.68 &60.52 &79.68 &50.88 &63.57 &38.48 \\
\hline
NoT & 75.19 & 46.12 & 81.60 & 51.03 & 86.54 & \textbf{87.89} & 69.12 & 79.84 \\
ToT & 81.98 & 58.72 & 77.81 & 51.24 & 91.04 & 79.26 & 75.32 & 82.52\\
GoT & 74.86 & 56.08 & \textbf{84.05} & 50.83 & 90.95 & \underline{84.57} & 66.14 & 81.02 \\
SCP & 81.71 & 60.42 & 80.93 & 52.70 & 91.22 & 84.32 & 72.35 &\underline{84.29}\\
CLEAR & 82.71 & 55.57 & 79.71 & 53.95 & \textbf{93.27} & 84.00 & 78.81 & \textbf{84.48} \\
\hline
Self Ask & 78.52 & 45.43 & 79.15 & 64.66 & 81.42 & 80.15 & 70.67 & 63.48 \\
Plan \& Solve & 81.72 & 39.51 & 67.56 & 66.32 & 90.60 & 81.83 & 77.00 & 62.63 \\
Self Discover & 80.32 & 59.42 & 78.93 & 65.49 & 91.35 & 81.16 & 74.81 & 80.43 \\
\hline
SEAR &81.45 &\underline{60.18} &79.71 &65.90 &90.02 &82.87 &\underline{80.23} &81.15 \\
SEAR\_U &82.18 &\textbf{61.75} &82.61 &\textbf{68.71} &\ \underline{92.78} &79.84 &\textbf{81.52} &82.00 \\
\hline
SEAR+R &\underline{82.71} &58.54 &81.05 &65.49 &89.39 &84.20 &78.04 &65.90 \\ 
SEAR\_U+R &\textbf{83.38} &56.58 & \underline{82.83} &\underline{67.36} &91.53 &85.52 &77.91 &67.08 \\
\bottomrule
\end{tabular} 
\caption{HCS scores (in \%) using Gemini 1.5 Flash, R stands for "Refactoring" and U stands for "Unified".Bold represents the best performer and the underlined represents the second best performer.}\label{tab:hcs_gemini}
\end{table}
\vspace{-8pt}
\begin{table}[!htbp]
\centering
\scriptsize
\setlength{\tabcolsep}{1.7pt}
\begin{tabular}{lrrrrrrrrr}\toprule
&wiki &multi &hitab &finqa &tatqa &fetaqa &squall &hybridqa \\\midrule
COT &78.92 &57.97 &77.59 &64.14 &92.91 &84.13 &67.57 &78.21 \\
F-COT &71.61 &55.32 &71.35 &64.97 &91.04 &77.81 &56.46 &34.62 \\
Decomp &79.79 &57.03 &76.14 &65.18 &92.65 &78.45 &62.40 &77.68 \\
EE &80.12 &56.77 &79.38 &56.03 &92.81 &83.88 &66.67 &79.58 \\
POT &79.59 &57.91 &76.25 &56.13 &90.15 &72.00 &72.35 &61.98 \\
\hline
NoT & 65.82 & 44.54 & \textbf{80.82} & 50.41 & 88.01 & \textbf{85.46} & 52.58 & 76.83 \\
ToT & 81.91 & 56.89 & 79.04 & 55.40 &\textbf{96.60} & 82.30 & 66.67 & \underline{80.49}\\
GoT & 71.54 & 52.04 & 74.58 & 51.35 & 90.90 & 81.68 & 53.61 & 75.58 \\
SCP & 79.05 & 57.59 & 79.71 & 55.19 & 92.29 & 84.19 & 66.53 &80.01\\
CLEAR & \underline{82.84} & 58.09 & 78.26 & 55.92 & 85.22 & 84.00 & 68.08 & 82.26 \\
\hline
Self Ask & 78.66 & 54.38 & 79.60 & 66.11 & 90.76 & 83.03 & 72.09 & 63.48\\
Plan \& Solve & 82.65 & 56.77 & 78.26 & 64.97 & 90.34 & 83.92 & 77.26 & 62.63\\
Self Discover & 82.71 & 56.46 & 79.60 & 65.70 & 91.67 & 84.51 & 70.28 & \textbf{80.43}\\
\hline
SEAR &80.19 &57.40 &77.37 &67.26 &92.42 &83.38 &69.64 &75.33 \\
SEAR\_U &79.92 &\textbf{61.00} &78.93 &\textbf{71.10} & \underline{92.91} &\underline{84.89} &76.74 &78.27 \\
\hline
SEAR + R &82.91 &56.65 &78.82 &66.94 &91.84 &84.77 &\textbf{79.33} &68.72 \\
SEAR\_U + R &\textbf{84.18} &\underline{59.29} &\underline{80.27} &\underline{69.75} &91.44 &84.39 &\underline{79.20} &70.48 \\
\bottomrule
\end{tabular}
\caption{HCS scores (in \%) using GPT 4o mini, R stands for "Refactoring" and U stands for "Unified".Bold represents the best performer and the underlined represents the second best performer.}\label{tab:hcs_gpt}
\end{table}
\vspace{-10pt}

\paragraph{Is table refactoring lossless?} While LLM-based refactoring may introduce a risk of hallucination, we empirically evaluate this using the AutoQA metric \cite{jain-etal-2024-structsum}, which measures answer accuracy on both original and refactored tables. As shown in Table~\ref{tab:autoQA_acc_table_refactoring}, the loss in fidelity is minimal. The slight drop in accuracy is primarily due to purposeful modifications, such as the addition of numerical units, adjustments to headers, and revised table titles. Although these changes alter the structure, they improve semantic clarity and enhance the tables’ utility for downstream reasoning tasks.

\begin{table}[!htbp]\centering
\scriptsize
\setlength{\tabcolsep}{1.7pt}

\begin{tabular}{lrrrrrrrrr}
\toprule
&wiki &multi &hitab &finqa &tatqa &fetaqa &squall &hybridqa \\\midrule
COT &81.05 &57.59 &82.95 &66.22 &91.00 &86.03 &75.45 &81.66 \\
F-COT &66.22 &39.82 &64.55 &51.77 &45.12 &52.78 &61.11 &33.31 \\
Decomp &82.85 &59.29 &81.84 &65.28 &93.18 &84.51 &73.51 &80.53 \\
EE &81.91 &58.92 &82.84 &61.75 &\underline{92.54} &86.62 &80.10 &81.07 \\
POT &76.53 &58.98 &67.56 &66.42 &91.40 &50.44 &68.22 &37.76 \\
\hline
NoT & 55.57 & 39.76 & 49.83 & 42.23 & 48.57 & 61.18 & 44.85 & 65.32 \\
ToT & \underline{84.57} & 45.35 & 74.99 & 57.58 & 82.67 & 83.50 & 78.29 & \underline{83.18}\\
GoT & 71.27 & 52.61 & 68.45 & 40.24 & 72.73 & \textbf{88.49} & 59.19 & 74.80 \\
SCP & 82.96 & 57.80 & 79.38 & 52.52 & 85.22 & 85.46 & 74.96 &79.75\\
CLEAR & 86.23 & 54.93 & 76.39 & 56.23 & 92.15 & 86.97 & 79.84 & 79.71 \\
\hline
Self Ask & 81.98 & 56.84 & 82.06 & 67.46 & 91.69 & 85.98 & 76.10 & 72.32\\
Plan \& Solve & 82.65 & 55.95 & 80.39 & 66.57 & 92.51 & 83.96 & 76.23 & 70.55\\
Self Discover & \textbf{85.77} & 57.91 & \textbf{83.95} & 66.11 & 92.87 & 86.09 & 79.33 & \textbf{83.25} \\
\hline
SEAR &82.65 &\underline{59.61} &\underline{83.05} &66.63 &92.34 &85.52 &81.40 &79.78 \\
SEAR\_U &82.05 &\textbf{62.19} &82.39 &\textbf{70.17} &\textbf{93.27} &87.04 &8\underline{82.04} &80.27 \\
\hline
SEAR + R &82.65 &57.09 &82.39 &67.26 &91.67 &86.85 &76.87 &67.74 \\
SEAR\_U + R & 85.11  &58.16 &83.05 &\underline{69.67} &92.89 &\underline{87.23} &\textbf{82.49} &72.16 \\
\bottomrule
\end{tabular}
\caption{HCS scores (in \%) using Llama 3.1 70B, R stands for "Refactoring" and U stands for "Unified".Bold represents the best performer and the underlined represents the second best performer.}\label{tab:hcs_llama}
\end{table}
\vspace{-10pt}

\begin{table}[!htp]\centering
\scriptsize
\setlength{\tabcolsep}{2.5pt}
\begin{tabular}{lrrrrrrrr}
\toprule
\textbf{Dataset} & \textbf{fetaqa} & \textbf{finqa} & \textbf{hitab}  & \textbf{multi} & \textbf{squall} & \textbf{tatqa} & \textbf{wiki} & \textbf{hybridqa} \\\midrule
\bf Accuracy & 99.41 & 95.36 & 98.06 & 88.04 & 86.66 & 99.40 & 96.43 & 84.59 \\\bottomrule
\end{tabular}
\vspace{-0.75em}
\caption{\small AutoQA Accuracy after refactoring Tables.}
\label{tab:autoQA_acc_table_refactoring}
\vspace{-1.5em}
\end{table}
\begin{figure}[t]
  \centering
  \includegraphics[width=\linewidth]{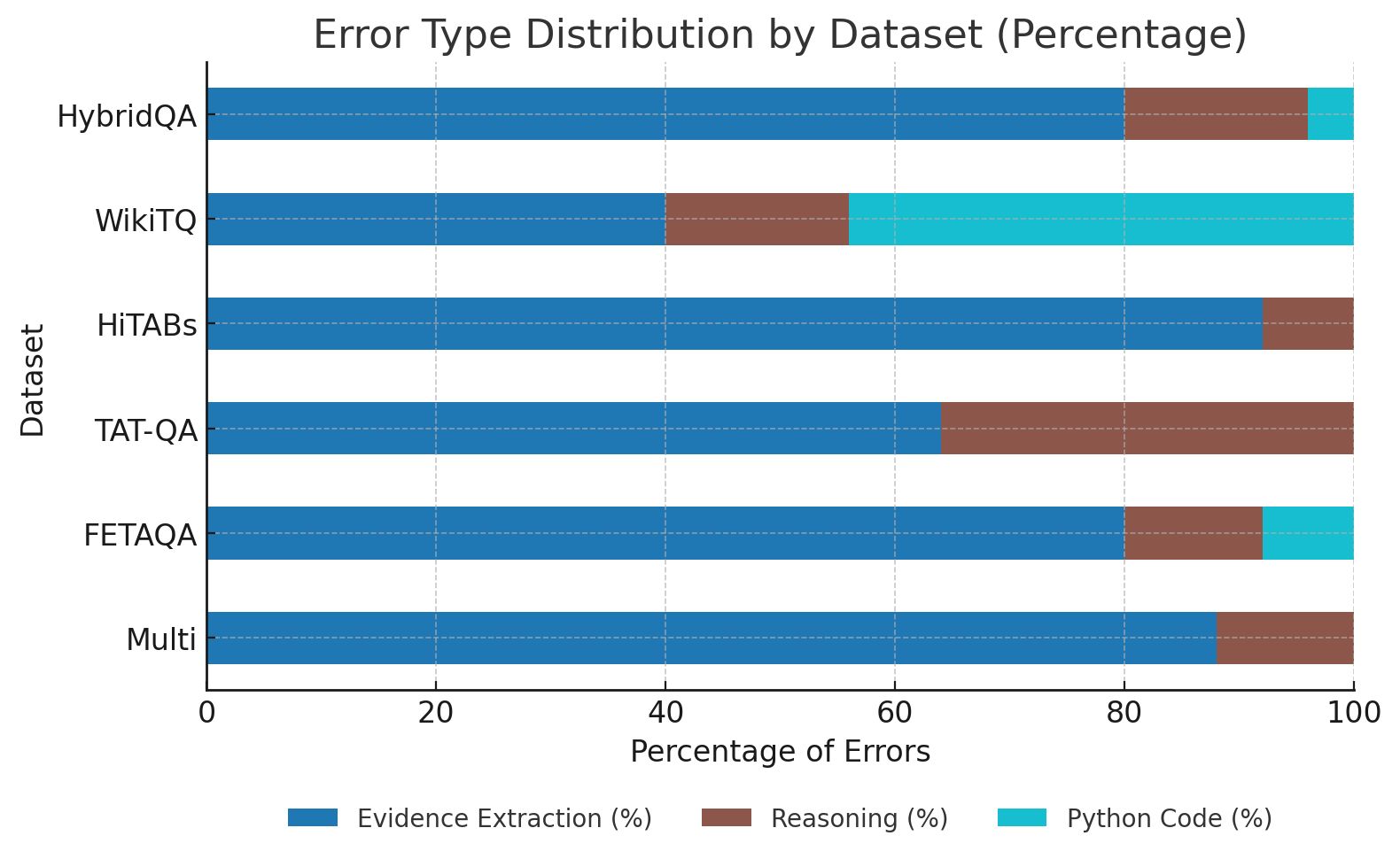}
  \caption{Distribution of error types (evidence extraction, reasoning, and Python code execution) across six benchmark datasets. Evidence extraction emerged as the dominant failure mode in five of the six cases.}
  \label{fig:error-dist}
\end{figure}

\paragraph{Error Analysis Summary.}
We conduct a fine‑grained error analysis across six datasets  as shown in Figure \ref{fig:error-dist} and find that evidence extraction is the most common failure mode, accounting for the majority of errors in five out of six cases. These errors arise from shallow string matching, ambiguous headers, and missed qualifiers (e.g., years, units, footnotes), leading models to anchor to plausible but incorrect cells, often before any reasoning or computation can take place. Reasoning errors are more prominent in datasets requiring temporal alignment or multi‑hop inference, such as TAT‑QA, while code‑generation failures dominate in WikiTQ due to parsing issues and faulty aggregation over semi‑structured tables. Overall, this suggests that early‑stage grounding remains the key bottleneck across tasks, with dataset‑specific challenges emerging in reasoning and execution stages. \textbf{Please refer to Appendix~\ref{sec:error-analysis-details}} for a detailed breakdown of error types by dataset.

\section{Discussion}

The Adaptive Framework consistently generalizes across multiple datasets by dynamically selecting appropriate reasoning paths. Table \ref{tab:reasoning_paths} summarizes the reasoning paths chosen by GPT-4o-mini, showing that Evidence Extraction is always included. This step helps the model focus on relevant information, aligning with human intuition (Section \ref{sec:adaptive_prompting}). For lookup-based questions, Evidence Extraction alone suffices, while more complex tasks require a combination of reasoning methods.

Datasets with long-form answers, such as FeTaQA, textual strategies works best. As shown in Table \ref{tab:hcs_llama}, for LLaMA 3.1 70B, FeTaQA achieves higher accuracy with CoT (84.13\%) and Decomposed Prompting (78.45\%). This trend is further supported by Table \ref{tab:reasoning_paths}, where Evidence Extraction + Decomposed Prompting is the most frequently chosen. Table \ref{tab:reasoning_methods} reinforces this, showing that 87\% of FeTaQA’s reasoning paths rely on textual methods, highlighting their effectiveness for free-form responses.

FinQA, which is heavy on numerical computation, favors symbolic methods. As seen in Table \ref{tab:hcs_llama}, PoT achieves the best performance, with F-CoT also performing well. Table \ref{tab:reasoning_paths} further confirms this, with Evidence Extraction + F-CoT as the most common reasoning path. Similarly, Table \ref{tab:reasoning_methods} shows that 88.25\% of FinQA’s reasoning paths involve PoT and F-CoT, emphasizing the strength of symbolic reasoning for computation-heavy datasets.

This pattern extends across datasets, with chosen reasoning paths aligning with their respective strengths. Table \ref{tab:reasoning_paths_llama} and \ref{tab:reasoning_path_gemini} in Appendix \ref{sec:appendix_rems_cae} provide reasoning path analysis for LLaMA 3.1 70B and Gemini-1.5-flash, respectively. By dynamically selecting the most effective reasoning approach based on question type and tabular context, the Adaptive Framework consistently delivers strong performance across diverse table structures and reasoning tasks.

\textbf{Impact of Table Refactoring.} Refactoring tabular data enhances LLM accuracy by improving clarity, structure, and accessibility. Table \ref{tab:dataset_evaluation} categorizes key refactoring techniques that aid model interpretation. In \emph{‘Table Structure’}, standardizing tables to Markdown format significantly improves performance. For instance, the Squall dataset, originally in JSON, benefits from this transformation. As shown in Table \ref{tab:hcs_gpt}, GPT-4o-mini with SEAR + Refactoring (79.33\%) outperforms SEAR (69.64\%) by 9.69\%, and SEAR\_U + Refactoring (79.20\%) exceeds SEAR\_U (76.74\%) by 2.46\%. Similarly, LLaMA 3.1 70B achieves its highest accuracy (82.49\%) with SEAR\_U + Refactoring. In \emph{‘Title Clarity’}, refining ambiguous or missing table titles improves context.

Figure \ref{fig:refactoring_prompt} illustrates how adding a player’s name in the title enhances model comprehension. \emph{‘Column/Row Headers’} are refined to eliminate ambiguity and better align entities. \emph{‘Data Formatting’} reduces redundant details, such as excessive decimal places, which can increase hallucinations as context size grows \cite{liu2023lostmiddlelanguagemodels}. Limiting decimals helps models focus and improves accuracy. \emph{‘Bolding and Emphasis’} highlights key details, directing the model's attention to relevant content. \emph{‘Other’} refinements, such as adding units, removing whitespace, and reformatting text, further enhance readability. The prompt for table refactoring is shown in Figure \ref{fig:prompt_table_refactoring}. 
\vspace{-5pt}
\begin{table}[!htbp]\centering
\scriptsize
\setlength{\tabcolsep}{1.3pt}
\begin{tabular}{llrrrrrrrr}\toprule
\multirow{2}{*}{\textbf{Reasoning Path}} & \multicolumn{8}{c}{\textbf{Datasets}} \\\cmidrule{2-9}
 & \textbf{fetaqa} & \textbf{finqa} & \textbf{hitab} & \textbf{hybridqa} & \textbf{multi} & \textbf{squall} & \textbf{tatqa} & \textbf{wiki} \\\midrule
\textbf{EE} & 175 & 46 & 476 &\bf 1332 & 194 & 13 &\bf 929 &\bf 703 \\
\textbf{EE,Decomp} &\bf 1365 & 65 & 191 & 28 & 127 & 160 & 249 & 293 \\
\textbf{EE,F-COT} & 23 &\bf 703 & 111 & 5 &\bf 335 &\bf 581 &\bf 547 & 246 \\
\textbf{EE,POT} & 9 & 138 & 107 & 143 & \bf 909 & 14 &\bf  482 & 186 \\
\textbf{COT,EE} & 1 & 1 & 4 & 12 & 5 & - & 5 & 32 \\
\textbf{COT,EE,Decomp} & 8 & 1 & 3 & 2 & - & 1 & 1 & 13 \\
\textbf{COT,EE,F-COT} & 1 & 7 & 1 & - & 5 & 5 & 12 & 17 \\
\textbf{COT,EE,POT} & - & 1 & 4 & 6 & 12 & - & 19 & 14 \\\midrule
\textbf{Total} & 1582 & 962 & 897 & 1528 & 1587 & 774 & 2244 & 1504 \\\bottomrule
\end{tabular}
\vspace{-0.75em}
\caption{\small Reasoning Path distribution for GPT-4o-mini.}
\label{tab:reasoning_paths}
\vspace{-1.5em}
\end{table}

\vspace{-0.25em}
\begin{table}[!htbp]\centering
\scriptsize
\setlength{\tabcolsep}{2.9pt}
\begin{tabular}{l|rr|rr|rr|rr|rr}
\hline
\textbf{Dataset} & \multicolumn{2}{c|}{\textbf{COT}} & \multicolumn{2}{c|}{\textbf{EE}} & \multicolumn{2}{c|}{\textbf{Decomp}} & \multicolumn{2}{c|}{\textbf{POT}} & \multicolumn{2}{c}{\textbf{F-COT}} \\
 & \textbf{\#} & \textbf{\%} & \textbf{\#} & \textbf{\%} & \textbf{\#} & \textbf{\%} & \textbf{\#} & \textbf{\%} & \textbf{\#} & \textbf{\%} \\\hline
\textbf{fetaqa} & 10 & 0.63 & 1582 & 100 & \textbf{1373} & \textbf{86.79} & 9 & 0.57 & 24 & 1.52 \\
\textbf{finqa} & 10 & 1.03 & 962 & 100 & 66 & 6.86 & 139 & 14.45 & \textbf{710} & \textbf{73.8} \\
\textbf{hitab} & 12 & 1.34 & 897 & 100 & \textbf{194} & \textbf{21.63} & 111 & 12.38 & 112 & 12.49 \\
\textbf{hybridqa} & 20 & 1.31 & 1528 & 100 & 30 & 1.96 & \textbf{149} & \textbf{9.75} & 5 & 0.33 \\
\textbf{multi} & 22 & 1.39 & 1587 & 100 & 127 & 8.01 & \textbf{921} & \textbf{58.03} & 132 & 8.32 \\
\textbf{squall} & 6 & 0.78 & 774 & 100 & 161 & 20.8 & 14 & 1.81 & \textbf{586} & \textbf{75.71} \\
\textbf{tatqa} & 37 & 1.65 & 2244 & 100 & 250 & 11.14 & 501 & 22.33 & \textbf{559} & \textbf{24.91} \\
\textbf{wiki} & 76 & 5.05 & 1504 & 100 & \textbf{306} & \textbf{20.35} & 200 & 13.3 & 263 & 17.49 \\
\bottomrule
\end{tabular}
\vspace{-0.75em}
\caption{\small Distribution of reasoning methods across all the datasets for GPT-4o-mini.}
\label{tab:reasoning_methods}
\vspace{-1.5em}
\end{table}
\vspace{-8pt}
\paragraph{SEAR in the Context of Agentic Frameworks.}
Agentic frameworks have gained attention for their ability to handle complex reasoning tasks through modular, interacting components such as planning, memory retrieval, and tool use. Although SEAR is not agentic by design, its structured reasoning process aligns with the modular philosophy of agentic systems. Each SEAR module could be instantiated as an individual agent within such a framework. However, the goal of this work is to explore how far prompting alone without external tools or orchestration can be used to address temporal table QA. This design choice prioritizes simplicity and self-containment. Importantly, the central challenge SEAR addresses is selecting and sequencing the appropriate reasoning strategies for a given question and table structure remains critical even within agentic systems. 

While agentic architectures offer a more general execution framework, they still depend on effective strategy selection. In this sense, SEAR provides a complementary perspective, offering insights into reasoning decomposition that could inform or enhance agent-based designs.

\section{Related Work}

\textbf{Tabular Reasoning.} LLMs have been widely applied to tabular reasoning tasks such as question answering, semantic parsing, and table-to-text generation \cite{chen2020tabfactlargescaledatasettablebased, gupta-etal-2020-infotabs, Zhang:2020:SET, Zhang:2020:WTE}. Early approaches like TAPAS \cite{Herzig_2020}, TaBERT \cite{yin2020tabert}, and TABBIE \cite{iida2021tabbie} improve table comprehension by integrating tabular and textual embeddings, allowing models to better process structured information. Other methods, such as Table2Vec \cite{Zhang_2019} and TabGCN \cite{pramanick2021joint}, explore alternative tabular representations, enhancing LLMs’ ability to infer relationships between table elements. However, these methods primarily focus on structured tables and do not explicitly address temporal reasoning, which introduces additional complexity when reasoning over tabular data.

\textbf{Symbolic Reasoning for Tables.} Recent work has explored symbolic reasoning for structured tables with predefined schemas, improving logical inference and data consistency \cite{cheng2023bindinglanguagemodelssymbolic, ye2023largelanguagemodelsversatile, wang2024chainoftableevolvingtablesreasoning}. These methods rely on well-defined structures to extract and process information effectively. However, they struggle with semi-structured and hierarchical tables, where relationships between data points are implicit rather than explicitly defined. 

\textbf{Other Reasoning Frameworks.} C.L.E.A.R \cite{deng2024enhancingtemporalunderstandingllms} demonstrated strong temporal reasoning on domain-specific semi-structured tables by integrating domain knowledge into responses.  Similarly, Meta-Reasoning Prompting (MRP)\cite{gao2024metareasoninglargelanguage} selects the optimal reasoning strategy through a two-step process but does not combine reasoning techniques for complex tasks. In contrast, our approach integrates both textual and symbolic reasoning to enhance performance across diverse table types while dynamically selecting the best reasoning path. Moreover, our SEAR-Unified prompt streamlines this into a single-step process, ensuring efficiency and consistency across different table structures.

\section{Conclusion and Future Work}
This paper introduces SEAR, an adaptive reasoning strategy for LLMs to tackle TTQA tasks, along with its consolidated version, SEAR\_Unified. Additionally, we take a step toward a unified table representation by incorporating table refactoring as an enhancement. Our study provides a comprehensive analysis of various reasoning strategies across eight diverse datasets, benchmarking SEAR and SEAR\_Unified against multiple baselines.

Results demonstrate that SEAR, SEAR\_Unified and with Table Refactoring significantly outperforms popular LLM reasoning methods, with SEAR\_Unified surpassing SEAR itself, showcasing its ability to optimize and streamline reasoning with minimal overhead. This highlights capability of modern LLMs to dynamically adjust reasoning within a single prompt, reducing the need for explicit multi-step processes. Our findings reinforce the importance of adaptive reasoning and structured table representation, paving the way for further advancements in LLM-based temporal table reasoning.

While SEAR-based approaches have significantly improved Temporal Table QA, several areas remain open for further exploration. In this work, we have explored Markdown as a unified tabular representation, exploring alternative formats such as JSON, CSV, or HTML may further improve adaptability across diverse table structures. Currently, our experiments relied on in-context learning, which can limit scalability and efficiency. Future work should explore lightweight adaptive reasoning techniques with self‑refinement loops, building on the flexibility demonstrated by SEAR. Lastly, valuating SEAR-based methods on additional domains, such as medical or scientific evolution datasets, would help validate the robustness of adaptive reasoning strategies for LLMs.

\section*{Limitations}
While our study has yielded interesting observations, it's crucial to acknowledge its limitations. A closer look at the HCS scores in Table \ref{tab:hcs_gemini}, \ref{tab:hcs_gpt}, \ref{tab:hcs_llama}, reveals that while improvements are observed for datasets with single table contexts, datasets containing multiple tables, such as MultiHierTT and Hybrid tables, show a decline in performance with SEAR-based approaches. This highlights a key limitation of our Table Refactoring method, suggesting that restructuring strategies may need further refinement to handle multi-table contexts effectively. Additionally, scalability remains a concern, as our approach relies on In-Context Learning (ICL), which may not scale effectively for large table datasets. The reliance on ICL-based reasoning can lead to performance bottlenecks.

\section*{Ethics Statement}
We confirm that our work adheres to the highest ethical standards in research and publication. We will publicly release our code and filtered datasets to enable the research community to validate and build upon our findings. We are committed to the responsible and fair use of computational linguistics methodologies. The claims in our paper accurately reflect the experimental results. While using black-box large language models introduces some stochasticity, we mitigate this by maintaining a fixed temperature. We utilize an AI assistive tools for writing while ensuring absence of bias. We provide comprehensive details on annotations, dataset splits, models used, and prompting methods tried, ensuring the reproducibility of our work.



\bibliography{custom}

\appendix




\section{Prompt Examples}
\label{sec:appendix_prompt_examples}
\begin{figure*}[h]
    \centering
    \includegraphics[width=0.9\textwidth]{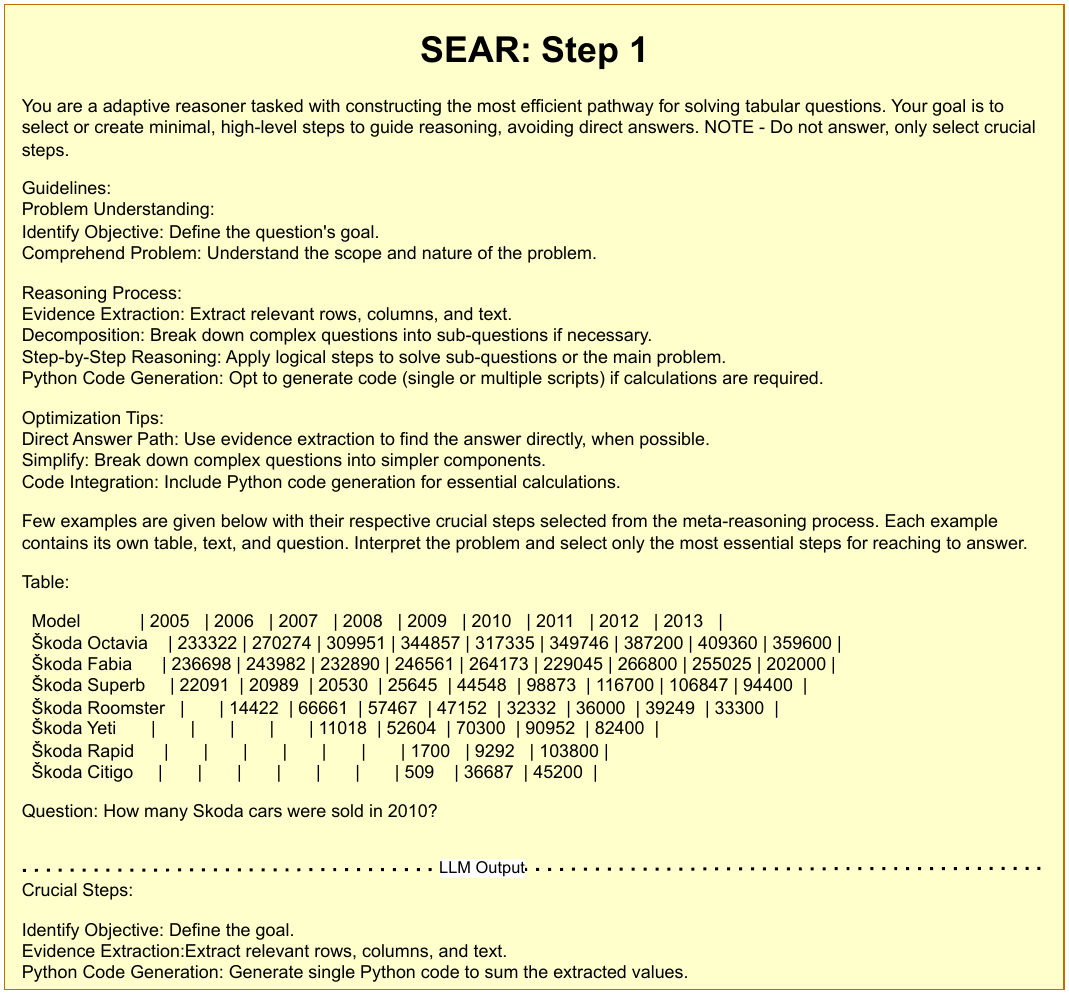} 
    \caption{\small Sear Step 1 Prompt Example}
    \label{fig:sears1_prompt}
\end{figure*}

\begin{figure*}[h]
    \centering
    \includegraphics[width=0.9\textwidth]{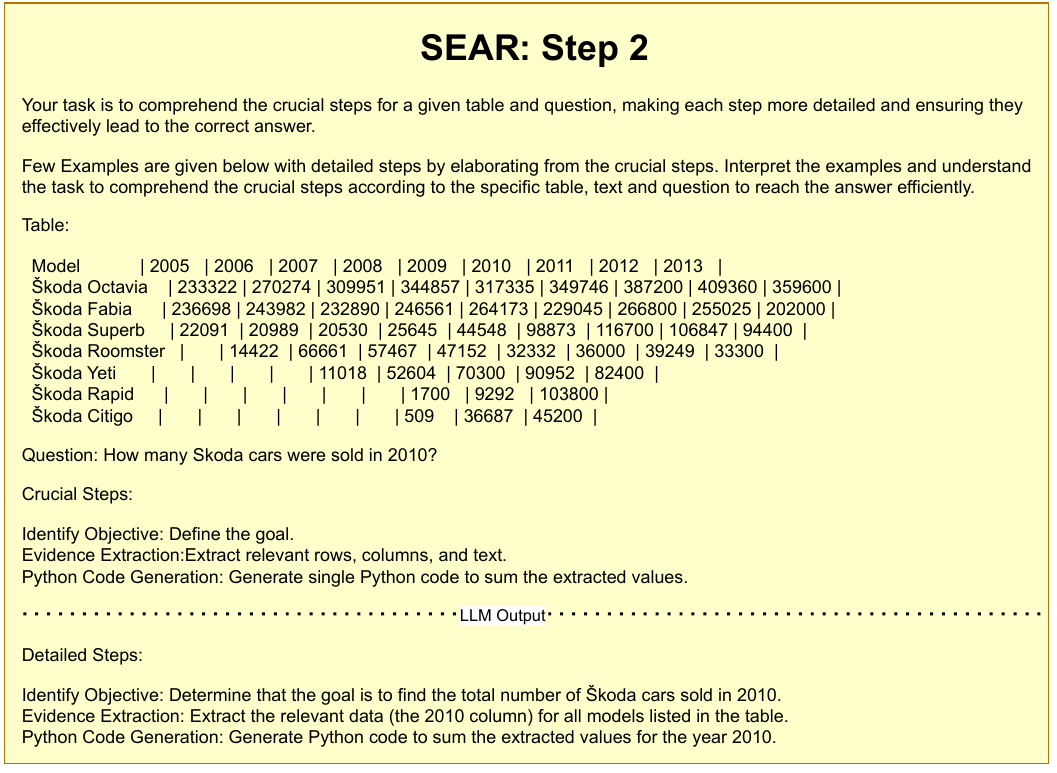} 
    \caption{\small Sear Step 2 Prompt Example}
    \label{fig:sears2_prompt}
\end{figure*}

\begin{figure*}[h]
    \centering
    \includegraphics[width=0.9\textwidth]{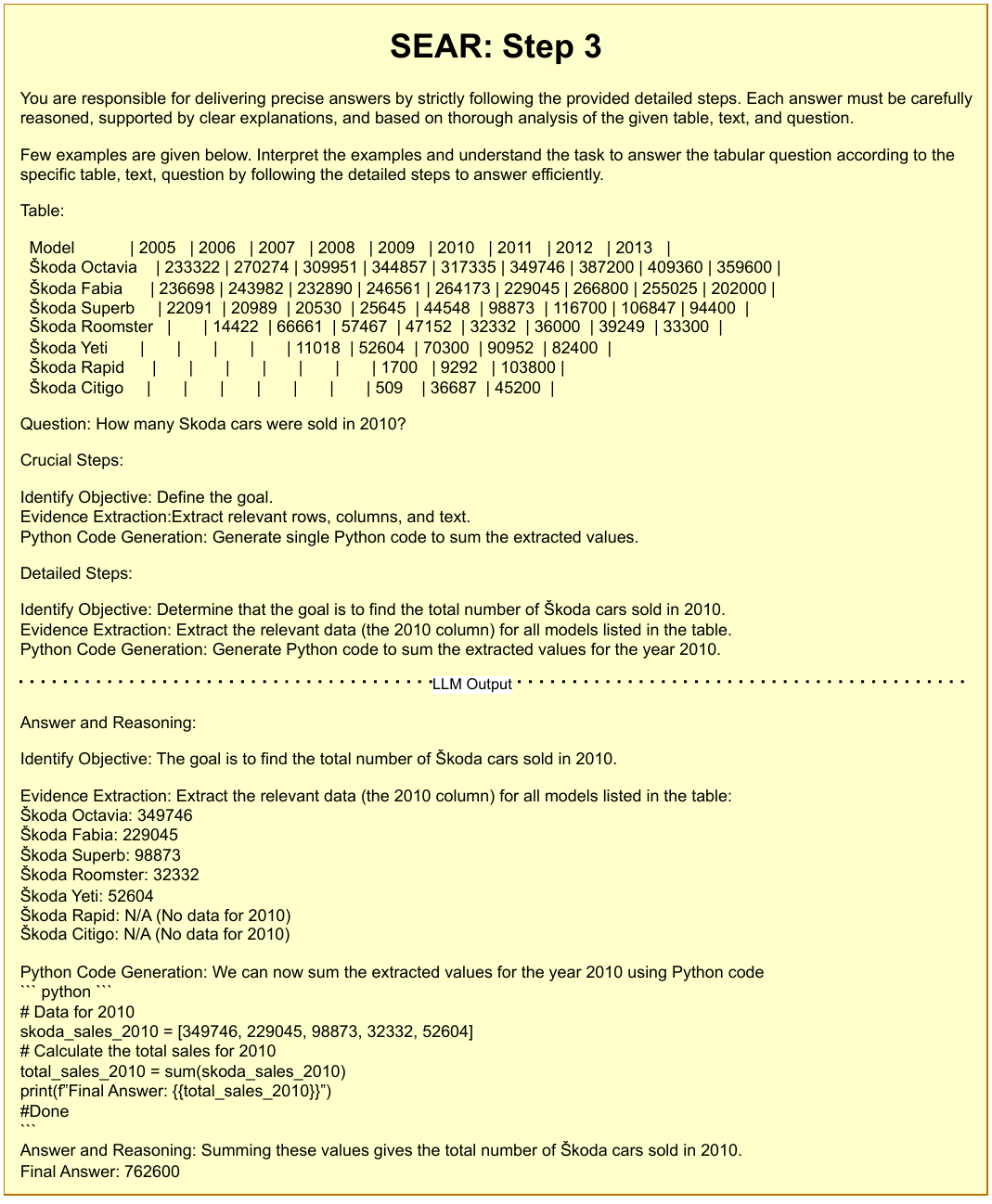} 
    \caption{\small Sear Step 3 Prompt Example}
    \label{fig:sears3_prompt}
\end{figure*}

\begin{figure*}[h]
    \centering
    \includegraphics[width=0.9\textwidth]{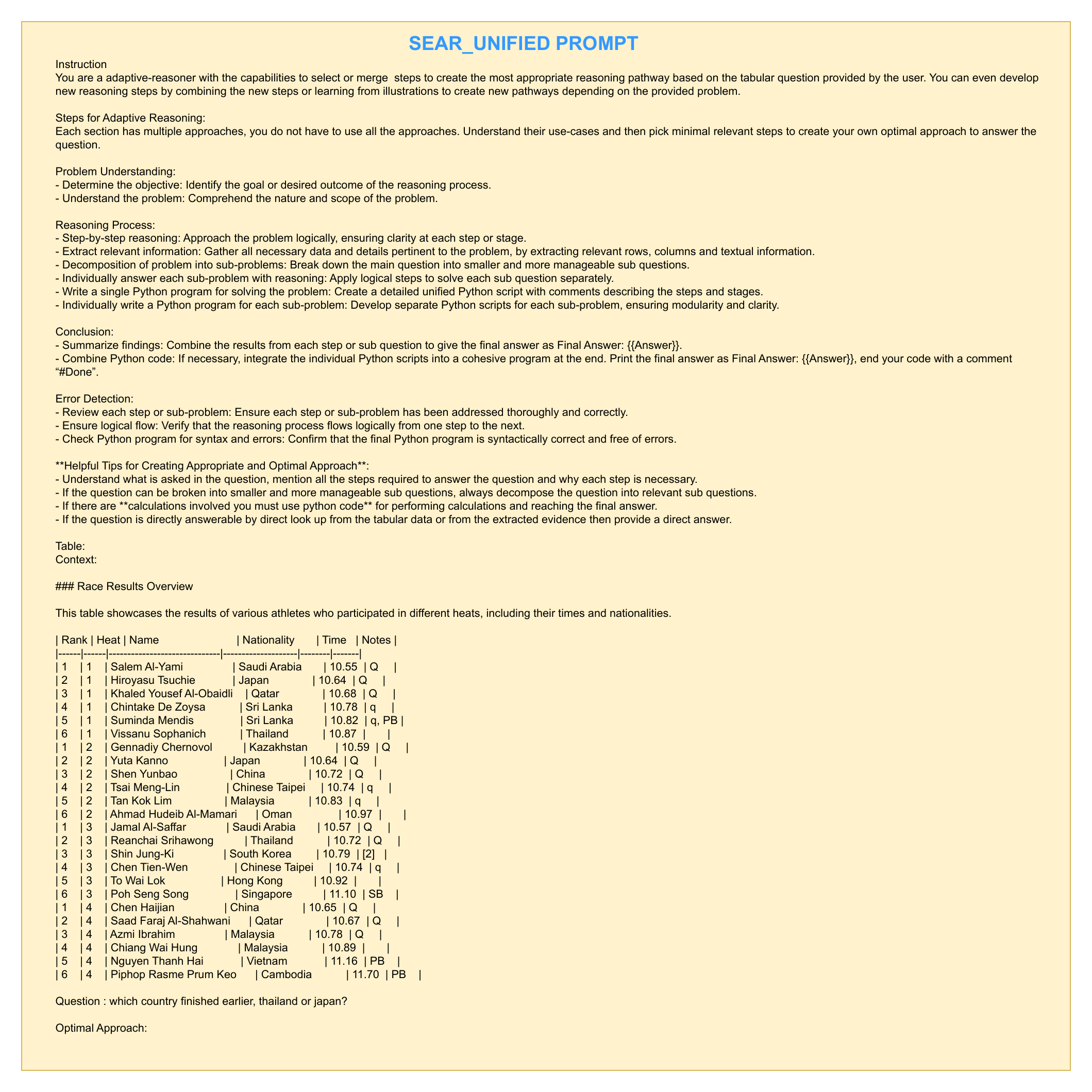} 
    \caption{\small SEAR\_UNIFIED Prompt Example}
    \label{fig:sear_prompt}
\end{figure*}

\begin{figure*}[h]
    \centering
    \includegraphics[width=0.9\textwidth]{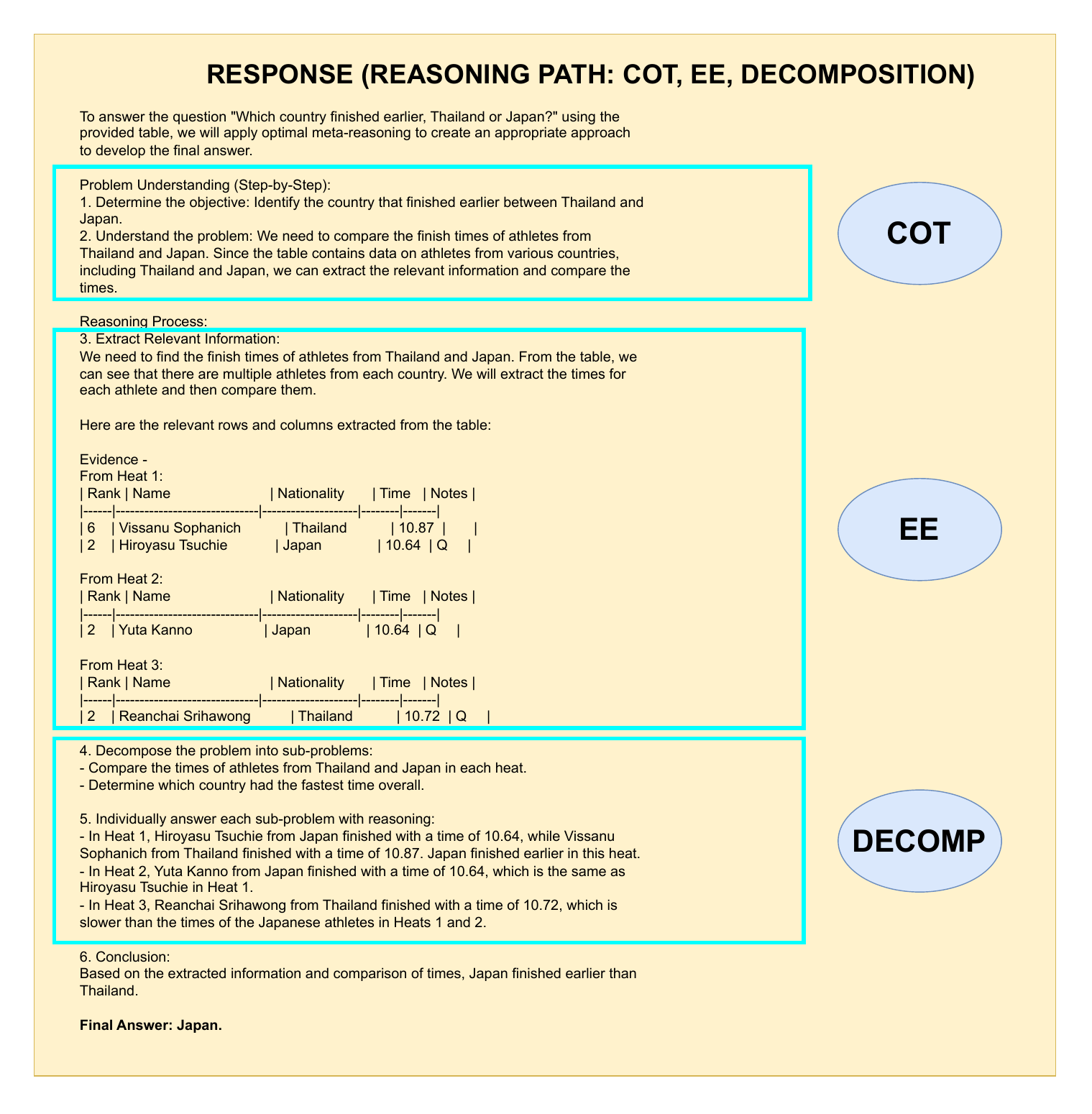} 
    \caption{\small The figure illustrates the response path followed by SEAR\_Unified Prompting. The reference prompt is provided in Figure \ref{fig:sear_prompt}}
    \label{fig:sear_response}
\end{figure*}

\begin{figure*}[htp]
    \centering
    \includegraphics[width=0.9\textwidth]{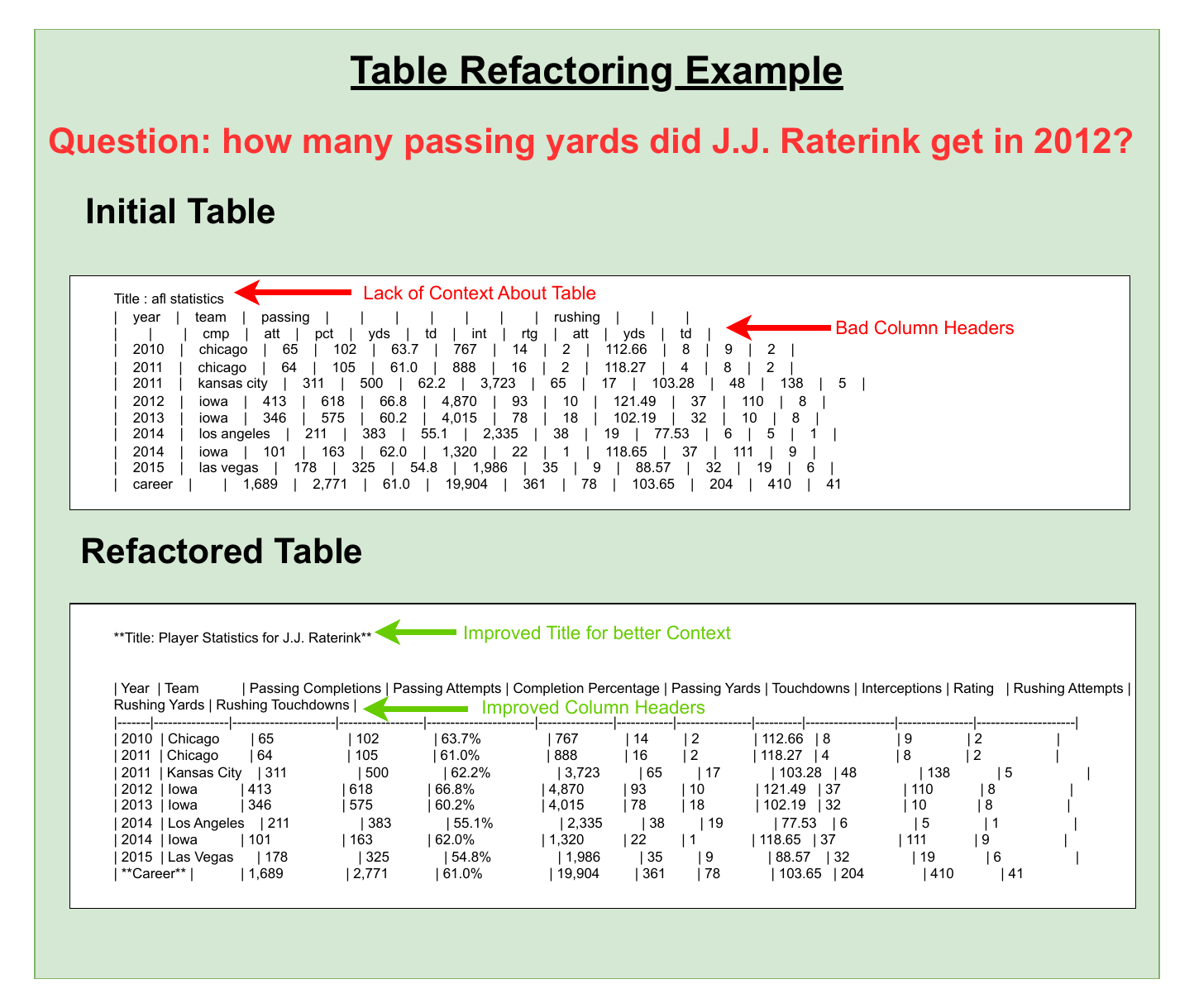} 
    \caption{\small Refactored Table Example}
    \label{fig:refactoring_prompt}
\end{figure*}

This section contains the Prompts example for standart 3-step SEAR (Figure \ref{fig:sears1_prompt}, \ref{fig:sears2_prompt}, \ref{fig:sears3_prompt}), SEAR Unified (Figure \ref{fig:sear_prompt} \ref{fig:sear_response}), Evaluation Prompt (Figure \ref{fig: prompt_llm_eval}) and Refactoring Prompt and Response (Figure \ref{fig:prompt_table_refactoring}, \ref{fig:refactoring_prompt}).

\begin{figure*}[!htb]
\centering
\begin{tcolorbox}[colback=white,arc=1mm] 
\begin{minipage}{\dimexpr\textwidth-2\fboxsep-2\fboxrule}
\centering
\raggedright
\textbf{Input :} \\

\vspace{\baselineskip}
\small
You are an expert LLM evaluator tasked with assessing the accuracy of model responses against gold standard answers. Your role is to determine if the core content and intent of the model's response align with the gold answer, considering various answer formats and implicit information.

\vspace{0.5em}
\textbf{Key Guidelines}

\begin{itemize}
    \item \textbf{Understand the question's essence}, including specific operations or units mentioned.
    \item \textbf{Compare model responses} to gold answers, focusing on key information.
    \item \textbf{Allow a small margin of error} (\(\pm 0.1\%\)) for numerical answers.
    \item \textbf{Recognize correct answers in different formats}, such as percentages and decimals.
    \item \textbf{Consider implicit information and context} in responses.
    \item \textbf{For list-type answers}:
    \begin{itemize}
        \item Evaluate based on content rather than order.
        \item If more than \textbf{two elements are missing} (context-dependent), evaluate as incorrect.
    \end{itemize}
    \item \textbf{Assess mathematical answers} based on value range unless a specific value is required.
    \item \textbf{Check for appropriate units} in mathematical answers.
\end{itemize}

\textbf{Final Judgment}

Provide a \textbf{"Yes" or "No"} judgment without explanation unless explicitly requested.

\end{minipage}
\end{tcolorbox}
\caption{Prompt for Contexual Answer Evaluation(CAV)}
\label{fig: prompt_llm_eval}
\end{figure*}

\begin{figure*}[!htb]
\centering
\begin{tcolorbox}[colback=white,arc=1mm] 
\begin{minipage}{\dimexpr\textwidth-2\fboxsep-2\fboxrule}
\centering
\raggedright
\textbf{Input :} \\

\vspace{\baselineskip}
\small

\textbf{Instruction}

You are given the following \textbf{Question} and \textbf{Context}. The \textbf{Context} includes a table that may be incomplete, ambiguous, or poorly structured. Your task is to produce a \textbf{cleaned version of the table} that improves its clarity and structure so that it can be correctly used to answer the \textbf{Question}.

\vspace{0.5em}
\textbf{Guidelines}

\begin{enumerate}
    \item \textbf{Do not add, remove, or alter any data}. Only restructure and clarify what is already present.
    
    \item You may improve the \textbf{table title} if it is missing or ambiguous:
    \begin{itemize}
        \item If a title is missing, infer an appropriate one based on the \textbf{question} and table content.
        \item If the existing title is unclear or misleading, revise it for clarity while keeping its original meaning.
    \end{itemize}
    
    \item You may improve the \textbf{table headers} if needed:
    \begin{itemize}
        \item Rename ambiguous column/row headers for clarity.
        \item Ensure column and row labels accurately describe their content.
    \end{itemize}
    
    \item You may fix \textbf{structural inconsistencies}:
    \begin{itemize}
        \item Align misaligned data properly under the correct headers.
        \item Ensure row and column structures are uniform.
        \item Remove redundant headers or merge split headers where necessary.
    \end{itemize}
    
    \item The data should be kept in the same order whenever possible. However, if \textbf{minor reordering of rows or columns} helps fix structural issues, you may do so \textbf{only if it does not change or omit any data}.
\end{enumerate}

\textbf{Output Format}

\begin{itemize}
    \item Provide only the \textbf{cleaned table} as your output in a structured format appropriate for the data in \textbf{Markdown format}.
    \item \textbf{Do not add any explanations, reasoning, or commentary}.
\end{itemize}

\hrulefill

\noindent
\textbf{Question}: \texttt{\{question\}}

\medskip

\noindent
\textbf{Context}:  
\texttt{\{context\}}

\medskip

\noindent
\textbf{Now produce just the cleaned table.}

\end{minipage}
\end{tcolorbox}
\caption{Prompt for Refactoring Tables.}
\label{fig:prompt_table_refactoring}
\end{figure*}

\section{REMS and CAE Results}
\label{sec:appendix_rems_cae}
This section contains the additional result for Reasoning Path Distribution for Llama and Gemini (Table \ref{tab:reasoning_paths_llama}, \ref{tab:reasoning_path_gemini}) and also contains complete result for CAE and REMS score for all GPT, Gemini and Llama (Table \ref{tab:rems_cae_gpt}, \ref{tab:rems_cae_gemini}, \ref{tab:rems_cae_llama}).\\

\textit{\underline{1. Relaxed Exact Match Score(REMS)}:} This
 metric uses an F1-score to measure token overlap between the predicted and gold answer, allowing partial matches for better precision-recall balance.
 Unlike strict exact match, REMS is more flexible with lexical variations. For numerical answers, it permits a ±5\% tolerance after decimal instead of
 token matching. For example, if the correct answer is 10.64, a prediction of 10.62 is accepted, while 11.64 is not.

 Despite its flexibility, REMS does not always reflect true semantic accuracy. High scores indicate strong token alignment, but valid paraphrases can be unfairly penalized. For instance, the correct answer “Barack Obama was the 44th President
 of the United States” would receive a high score for “Obama was the 44th U.S. President” due to token overlap, but “Obama, a politician, led the
 U.S.” may score lower despite being factually correct. This limitation makes careful interpretation\\


\textit{\underline{2. Contextual Answer Evaluation(CAE)}:} CAE is an LLM-based scoring method that assesses responses based on meaning rather than exact token overlap. Using a carefully crafted prompt, it determines whether a response correctly conveys the intended information. Unlike traditional lexical matching, CAE accounts for paraphrasing and rewording, ensuring a more nuanced assessment of correctness, particularly for complex or free-form answers. The full CAE prompt used for evaluation is provided in Figure \ref{fig: prompt_llm_eval}


\begin{table*}[!htp]\centering
\scriptsize
\setlength{\tabcolsep}{2.0pt}
\begin{tabular}{l|r|r|r|r|r|r|r|r}
\hline
\textbf{Reasoning Path} & \textbf{fetaqa} & \textbf{finqa} & \textbf{hitab} & \textbf{hybridqa} & \textbf{multi} & \textbf{squall} & \textbf{tatqa} & \textbf{wiki} \\\hline
\textbf{EE} & 221 & 39 & 561 & 1072 & 356 & 9 & 1040 & 987 \\
\textbf{EE,Decomp} & 553 & 21 & 19 & 8 & 33 & 28 & 59 & 81 \\
\textbf{EE,F-COT} & 571 & 853 & 123 & 35 & 262 & 709 & 391 & 236 \\
\textbf{EE,POT} & 234 & 45 & 194 & 405 & 919 & 25 & 753 & 187 \\
\textbf{COT,EE} & - & - & - & 6 & 5 & - & 1 & 7 \\
\textbf{COT,EE,Decomp} & 3 & - & - & 2 & 10 & 1 & - & 2 \\
\textbf{COT,EE,F-COT} & - & 3 & - & - & - & 2 & - & 4 \\
\textbf{POT} & - & 1 & - & - & 2 & - & - & - \\\hline
\textbf{Total} & 1582 & 962 & 897 & 1528 & 1587 & 774 & 2244 & 1504 \\\hline
\end{tabular}
\vspace{-0.75em}
\caption{\small Reasoning Path distribution across all datasets for Llama 3.1 70B.}
\label{tab:reasoning_paths_llama}
\vspace{-1.5em}
\end{table*}

\begin{table*}[!htp]\centering
\scriptsize
\setlength{\tabcolsep}{2.0pt}
\begin{tabular}{l|r|r|r|r|r|r|r|r}
\hline
\textbf{Reasoning Path} & \textbf{fetaqa} & \textbf{finqa} & \textbf{hitab} & \textbf{hybridqa} & \textbf{multi} & \textbf{squall} & \textbf{tatqa} & \textbf{wiki} \\\hline
\textbf{EE} & 982 & 106 & 675 & 1492 & 155 & 112 & 1160 & 875 \\
\textbf{EE,DecompE} & 197 & 16 & 6 & 2 & 87 & 17 & 9 & 186 \\
\textbf{EE,F-COT} & 175 & 796 & 29 & - & 333 & 516 & 49 & 173 \\
\textbf{EE,POT} & 191 & 42 & 186 & 33 & 1010 & 119 & 1025 & 268 \\
\textbf{COT,EE} & 25 & - & - & 1 & - & 1 & 1 & 2 \\
\textbf{COT,EE,Decomp} & 3 & - & - & - & - & 1 & - & - \\
\textbf{COT,EE,F-COT} & 2 & 1 & - & - & - & 6 & - & - \\
\textbf{COT,EE,POT} & 7 & - & 1 & - & - & 1 & - & - \\
\textbf{Decomp} & - & 1 & - & - & - & - & - & - \\
\textbf{POT} & - & - & - & - & 2 & 1 & - & - \\\hline
\textbf{Total} & 1582 & 962 & 897 & 1528 & 1587 & 774 & 2244 & 1504 \\\hline
\end{tabular}
\vspace{-0.75em}
\caption{\small Reasoning Path distribution across all datasets for Gemini-1.5-Flash.}
\label{tab:reasoning_path_gemini}
\vspace{-1.5em}
\end{table*}

\begin{table*}[!htp]
\centering
\hspace{-1.2cm}
\scriptsize
\setlength{\tabcolsep}{1.7pt}
\begin{tabular}{lrrrrrrrrrrrrrrrrr}\toprule
&\multicolumn{2}{c}{wiki} &\multicolumn{2}{c}{multi} &\multicolumn{2}{c}{hitab} &\multicolumn{2}{c}{finqa} &\multicolumn{2}{c}{tatqa} &\multicolumn{2}{c}{fetaqa} &\multicolumn{2}{c}{squall} &\multicolumn{2}{c}{hybridqa} \\\cmidrule{2-17}
&REMS &CAE &REMS &CAE &REMS &CAE &REMS &CAE &REMS &CAE &REMS &CAE &REMS &CAE &REMS &CAE \\\midrule
COT &77.31 &76.66 &\textbf{57.49} &49.72 &75.26 &74.58 &58.94 &57.07 &81.68 &87.48 &28.38 &84.13 &66.27 &65.25 &74.07 &76.51 \\
F\textminus COT &67.85 &67.82 &49.39 &51.35 &41.44 &69.79 &60.78 &61.12 &67.36 &86.76 &\textbf{40.46} &77.69 &52.97 &53.36 &29.78 &32.79 \\
Decomp &77.69 &76.60 &56.12 &49.02 &73.19 &73.36 &60.40 &58.21 &86.13 &87.25 &28.71 &78.45 &61.07 &59.30 &74.71 &74.87 \\
EE &78.57 &77.86 &56.32 &48.27 &\textbf{76.16} &76.92 &50.94 &46.88 &\textbf{90.22} &88.06 &28.42 &83.82 &65.55 &64.60 &\textbf{75.85} &\textbf{76.96} \\
POT &76.28 &75.93 &53.41 &53.12 &41.92 &73.47 &51.88 &52.49 &66.88 &86.10 &29.71 &72.00 &65.90 &69.12 &58.66 &60.27 \\
\hline
NoT &63.07 &63.56 &39.04 &38.44 &69.96 &76.25 &44.22 &46.05 &72.78 &82.58 &29.23 &85.46 &51.11 &50.52 &72.17 &75.65 \\
ToT &79.79 &78.92 &53.00 &52.43 &68.39 &76.92 &51.96 &49.90 &81.13 &88.01 &30.15 &82.17 &64.98 &63.44 &76.96 &78.01 \\
GoT &69.33 &67.09 &48.80 &45.05 &65.91 &71.68 &47.41 &46.36 &83.30 &86.14 &28.95 &81.67 &52.67 &48.58 &71.79 &72.05 \\
SCP &77.10 &76.73 &56.44 &52.11 &74.58 &77.15 &51.90 &50.00 &84.71 &86.63 &28.51 &84.13 &64.71 &64.47 &75.49 &78.73 \\
CLEAR &80.23 &79.72 &52.67 &57.40 &68.62 &75.81 & & &85.23 &91.13 &29.28 &83.94 &65.85 &66.28 &77.98 &79.84 \\
\hline
SEAR &78.32 &76.60 &54.70 &50.98 &67.36 &74.58 &62.52 &60.91 &81.94 &85.83 &29.53 &83.38 &67.56 &60.72 &72.07 &73.63 \\
SEAR\_U &77.50 &77.53 &56.39 &\textbf{56.84} &71.78 &76.70 &\textbf{62.87} &\textbf{67.57} &88.31 &\textbf{89.75} &31.06 &\textbf{84.89} &72.26 &73.77 &74.96 &75.85 \\
\hline
SEAR + R &80.51 &79.39 &54.04 &51.10 &68.40 &75.92 &61.88 &60.08 &81.63 &85.87 &29.71 &84.39 &\textbf{76.85} &74.03 &65.89 &66.03 \\
SEAR\_U + R &\textbf{81.14} &\textbf{81.25} &55.54 &55.51 &72.13 &\textbf{77.59} &62.43 &66.53 &86.56 &88.23 &30.47 &84.70 &76.21 &\textbf{76.87} &66.96 &67.74 \\
\bottomrule
\end{tabular}
\caption{REMS \& CAE score (in \%) for all reasoning strategies across all datasets using GPT‑4o mini. R stands for “Refactoring,” U for “Unified.”}
\label{tab:rems_cae_gpt}
\end{table*}

\begin{table*}[!htp]
\centering
\hspace{-1.2cm}
\scriptsize
\setlength{\tabcolsep}{1.7pt}
\begin{tabular}{lrrrrrrrrrrrrrrrrr}\toprule
&\multicolumn{2}{c}{wiki} &\multicolumn{2}{c}{multi} &\multicolumn{2}{c}{hitab} &\multicolumn{2}{c}{finqa} &\multicolumn{2}{c}{tatqa} &\multicolumn{2}{c}{fetaqa} &\multicolumn{2}{c}{squall} &\multicolumn{2}{c}{hybridqa} \\\cmidrule{2-17}
&REMS &CAE &REMS &CAE &REMS &CAE &REMS &CAE &REMS &CAE &REMS &CAE &REMS &CAE &REMS &CAE \\\midrule
COT &71.86 &71.28 &57.29 &39.26 &73.97 &74.25 &58.00 &39.29 &80.81 &85.34 &28.25 &71.24 &69.44 &69.66 &77.29 &76.57 \\
F-COT &64.76 &57.51 &58.36 &47.83 &35.68 &49.34 &60.60 &34.20 &64.86 &74.88 &\textbf{37.05} &55.69 &60.40 &60.73 &17.86 &15.97 \\
Decomp &76.26 &75.00 &58.90 &41.84 &71.70 &72.44 &60.72 &32.22 &84.23 &85.12 &29.91 &67.07 &66.01 &65.98 &72.94 &69.31 \\
EE &74.24 &72.81 &59.02 &42.41 &74.61 &76.43 &54.54 &30.46 &\textbf{86.14} &86.27 &28.63 &77.62 &71.89 &72.03 &74.12 &68.72 \\
POT &72.65 &66.69 &\textbf{60.00} &47.01 &41.37 &67.54 &54.90 &58.10 &66.74 &75.61 &26.73 &50.88 &62.66 &62.99 &38.18 &33.84 \\
\hline
NoT &70.40 &73.07 &40.59 &41.46 &72.08 &75.59 &58.32 &64.24 &69.41 &76.69 &30.73 &87.99 &65.03 &67.05 &75.43 &77.68 \\
ToT &79.79 &78.92 &54.65 &54.88 &70.33 &75.70 &41.68 &47.40 &76.84 &86.68 &29.48 &79.20 &71.90 &73.00 &79.42 &80.17 \\
GoT &72.82 &70.88 &51.20 &50.03 &68.41 &82.16 &48.34 &46.57 &79.15 &85.34 &29.57 &84.45 &62.08 &63.05 &77.14 &78.53 \\
SCP &79.40 &78.92 &57.72 &56.27 &75.90 &78.37 &46.53 &46.26 &81.38 &86.72 &27.85 &84.32 &68.82 &70.80 &79.26 &82.00 \\
CLEAR &79.82 &79.79 &56.08 &0.06 &70.38 &76.92 &49.47 &48.13 &79.94 &90.42 &28.63 &83.94 &75.32 &77.26 &79.59 &81.81 \\
\hline
SEAR &79.08 &78.19 &57.15 &54.69 &74.93 &76.81 &59.90 &61.02 &75.07 &83.87 &28.75 &82.87 &76.14 &68.60 &\textbf{77.61} &78.08 \\
SEAR\_U &79.32 &80.32 &59.27 &\textbf{57.34} &78.53 &79.38 &\textbf{63.16} &\textbf{65.59} &82.70 &\textbf{86.68} &31.57 &79.77 &\textbf{77.29} &\textbf{79.59} &77.11 &\textbf{79.84} \\
\hline
SEAR + R &80.27 &78.46 &55.32 &52.30 &75.08 &77.37 &59.88 &60.50 &73.57 &84.54 &28.97 &84.20 &76.13 &72.09 &62.43 &62.24 \\
SEAR\_U + R &\textbf{80.78} &\textbf{81.32} &53.09 &53.62 &\textbf{78.94} &\textbf{79.60} &61.98 &63.83 &82.20 &85.65 &32.89 &\textbf{85.52} &75.16 &75.97 &62.96 &64.86 \\
\bottomrule
\end{tabular}
\caption{REMS \& CAE score (in \%) for all reasoning strategies across all datasets using Gemini1.5 Flash. R stands for “Refactoring,” U for “Unified.”}
\label{tab:rems_cae_gemini}
\end{table*}

\begin{table*}[!htp]
\centering
\hspace{-1.2cm}
\scriptsize
\setlength{\tabcolsep}{1.7pt}
\begin{tabular}{lrrrrrrrrrrrrrrrrr}\toprule
&\multicolumn{2}{c}{wiki} &\multicolumn{2}{c}{multi} &\multicolumn{2}{c}{hitab} &\multicolumn{2}{c}{finqa} &\multicolumn{2}{c}{tatqa} &\multicolumn{2}{c}{fetaqa} &\multicolumn{2}{c}{squall} &\multicolumn{2}{c}{hybridqa} \\\cmidrule{2-17}
&REMS &CAE &REMS &CAE &REMS &CAE &REMS &CAE &REMS &CAE &REMS &CAE &REMS &CAE &REMS &CAE \\\midrule
COT &79.20 &78.86 &56.91 &48.71 &80.77 &\textbf{81.38} &60.91 &60.81 &83.69 &86.10 &28.07 &86.03 &73.21 &73.39 &79.10 &\textbf{79.78} \\
F-COT &63.02 &62.43 &37.21 &37.30 &37.35 &61.76 &48.14 &48.44 &59.67 &61.72 &25.34 &52.72 &56.53 &58.01 &30.30 &31.28 \\
Decomp &80.71 &80.78 &58.39 &52.24 &78.71 &80.72 &60.50 &59.77 &86.62 &86.41 &29.36 &84.51 &71.00 &71.58 &\textbf{79.98} &77.75 \\
EE &80.30 &79.79 &57.70 &48.27 &\textbf{81.42} &80.05 &57.03 &53.53 &89.09 &87.70 &28.63 &86.62 &78.12 &77.78 &78.33 &78.73 \\
POT &74.74 &73.34 &56.47 &55.14 &37.05 &65.44 &62.44 &61.75 &65.02 &87.17 &20.25 &50.44 &63.43 &64.73 &35.63 &35.60 \\
\hline
NoT &51.86 &52.39 &30.77 &34.85 &43.25 &46.82 &33.88 &39.50 &41.53 &46.75 &20.86 &61.19 &44.86 &47.03 &68.12 &69.50 \\
ToT &82.28 &81.72& 40.89 &46.06 &78.48 &80.71 &55.79 &49.06 &86.05 &90.01 &29.13 &83.44 &74.88 &75.97 &78.24 &80.96 \\
GoT &69.34 &68.02 &50.49 &48.08 &65.25 &66.33 &36.59 &40.24 &72.59 &77.58 &30.22 &88.50 &59.04 &61.88 &70.42 &73.23 \\
SCP &82.57 &85.10 &55.19 & 59.48 &80.15 &84.05 &52.65 &51.98 &84.19 &90.06 &28.68 &85.40 &77.03 &77.39 &77.15 &79.71 \\
CLEAR &83.49 &82.91 &83.50 &85.95 &83.50 &85.95 &36.50 &42.20 &90.06 &92.15 &29.36 &86.92 &77.39 &79.84 &75.58 &77.55 \\
\hline
SEAR &80.69 &78.79 &57.76 &50.79 &75.45 &78.60 &61.40 &60.40 &84.67 &\textbf{88.41} &29.47 &85.52 &78.74 &72.22 &76.43 &77.29 \\
SEAR\_U &78.91 &79.26 &\textbf{60.02} &\textbf{58.03} &75.12 &79.38 &\textbf{63.30} &\textbf{66.01} &\textbf{89.20} &86.36 &34.15 &87.04 &78.74 &80.62 &77.11 &78.24 \\
\hline
SEAR + R &80.17 &78.46 &54.97 &48.02 &75.77 &78.37 &62.00 &61.43 &81.71 &86.99 &29.53 &86.85 &73.95 &70.67 &67.35 &70.75 \\
SEAR\_U + R &\textbf{82.53} &\textbf{82.05} &56.15 &52.68 &76.19 &77.70 &61.66 &66.03 &86.58 &86.47 &\textbf{34.83} &\textbf{87.17} &\textbf{79.01} &\textbf{80.68} &67.11 &67.80 \\
\bottomrule
\end{tabular}
\caption{REMS \& CAE score (in \%) for all reasoning strategies across all datasets using Llama3.170B. R stands for “Refactoring,” U for “Unified.”}
\label{tab:rems_cae_llama}
\end{table*}

\section{Full Table and Context used in Figure \ref{fig:table_examples} }
\label{sec: appendix_fig1_example}

This section includes the actual table and context represented in Figure \ref{fig:table_examples}. FinQA Table \ref{tab:lease_payments_finqa}, FetaQA Table \ref{tab:filmograph_fetaqa}, WikiTabQA Table \ref{tab:sponsorship_history_wiki} and MultiHiertt Table \ref{tab:benefits_tab0_multi}, \ref{tab:loans_tab1_multi}.

\begin{table}[!htbp]
\small
    \centering
    \begin{tabular}{|l|r|r|r|}
        \hline
        \textbf{Benefit Plan} & \textbf{2017} & \textbf{2016} & \textbf{2015} \\
        \hline
        Pension Plan & 3856 & 3979 & 2732 \\
        Health Plan & 11426 & 11530 & 8736 \\
        Other plans & 1463 & 1583 & 5716 \\
        Total plan contributions & 16745 & 17092 & 17184 \\
        \hline
    \end{tabular}
    \caption{Benefit Plan Contributions, Benefits, MultiHiertt example Table 0}
    \label{tab:benefits_tab0_multi}
\end{table}

\begin{table*}[h]
\centering
\hspace{-1.2cm}
\scriptsize
\setlength{\tabcolsep}{1.7pt}
    \begin{tabular}{|c|c|c|c|c|}
        \hline
        \textbf{Year} & \textbf{Kit Manufacturer} & \textbf{Shirt Sponsor} & \textbf{Back of Shirt Sponsor} & \textbf{Short Sponsor} \\
        \hline
        1977–1978 & - & National Express & - & - \\
        1982–1985 & Umbro & - & - & - \\
        1985–1986 & Umbro & Whitbread & - & - \\
        1986–1988 & Henson & Duraflex & - & - \\
        1988–1989 & - & Gulf Oil & - & - \\
        1991–1993 & Technik & Gulf Oil & - & - \\
        1993–1994 & Club Sport & Gulf Oil & - & - \\
        1994–1995 & Klūb Sport & Empress & - & - \\
        1995–1996 & Matchwinner & Empress & - & - \\
        1996–1997 & UK & Endsleigh Insurance & - & - \\
        1997–1999 & Errea & Endsleigh Insurance & - & - \\
        1999–2004 & Errea & Towergate Insurance & - & - \\
        2004–2008 & Errea & Bence Building Merchants & - & - \\
        2008– & Errea & Mira Showers & - & - \\
        2009–2011 & Errea & Mira Showers & PSU Technology Group & - \\
        2011–2013 & Errea & Mira Showers & Barr Stadia & Gloucestershire Echo \\
        2013– & Errea & Mira Showers & Gloucestershire College & Gloucestershire Echo \\
        \hline
    \end{tabular}
    \caption{Historical Sponsorship and Kit Manufacturer Data, WikiTabQA example}
    \label{tab:sponsorship_history_wiki}
\end{table*}

\begin{table*}[h]
    \centering
    \resizebox{\textwidth}{!}{ 
    \begin{tabular}{|c|l|l|l|p{6cm}|}
        \hline
        \textbf{Year} & \textbf{Title} & \textbf{Role} & \textbf{Director} & \textbf{Notes} \\
        \hline
        2000 & The Apocalypse & Johanan & Raffaele Mertes & - \\
        2002 & Tom \& Thomas & Tom Sheppard / Thomas & Esmé Lammers & - \\
        2003 & Behind Closed Doors & Sam Goodwin & Louis Caulfield & - \\
        2003 & Shanghai Knights & Charlie Chaplin & David Dobkin & - \\
        2004 & Dead Cool & George & David Cohen & - \\
        2006 & The Thief Lord & Prosper & Richard Claus & - \\
        2006 & The Illusionist & Young Eisenheim & Neil Burger & - \\
        2006 & Fast Learners & Neil & Christoph Röhl & Short film \\
        2006 & The Best Man & Michael (Aged 15) & Stefan Schwartz & - \\
        2007 & The Magic Door & Flip & Paul Matthews & - \\
        2008 & Dummy & Danny & Matthew Thompson & Nominated — ALFS Award \\
        2008 & Angus, Thongs & Robbie Jennings & Gurinder Chadha & - \\
        2009 & The Greatest & Bennett Brewer & Shana Feste & - \\
        2009 & Nowhere Boy & John Lennon & Sam Taylor-Johnson & Empire Award for Best... \\
        2010 & Kick-Ass & David "Dave" Lizewski & Matthew Vaughn & Nominated — Empire Award... \\
        2010 & Chatroom & William Collins & Hideo Nakata & - \\
        2011 & Albert Nobbs & Joe Mackins & Rodrigo García & - \\
        2012 & Savages & Ben & Oliver Stone & - \\
        2012 & Anna Karenina & Count Vronsky & Joe Wright & Final time credited as... \\
        2013 & Kick-Ass 2 & David "Dave" Lizewski & Jeff Wadlow & First time credited as... \\
        2014 & Captain America: Winter Soldier & Pietro Maximoff & Anthony and Joe Russo & Uncredited cameo \\
        2014 & Godzilla & Lt. Ford Brody & Gareth Edwards & - \\
        2015 & Avengers: Age of Ultron & Pietro Maximoff & Joss Whedon & - \\
        2016 & Nocturnal Animals & Ray Marcus & Tom Ford & Golden Globe Award for... \\
        2017 & The Wall & Isaac & Doug Liman & - \\
        2018 & Outlaw King & James Douglas & David Mackenzie & - \\
        2018 & A Million Little Pieces & James Frey & Sam Taylor-Johnson & - \\
        2020 & Kingsman: The Great Game & - & Matthew Vaughn & Filming \\
        \hline
    \end{tabular}
    }
    \caption{Aaron Taylor-Johnson Filmography, example FeTaQA}
    \label{tab:filmograph_fetaqa}
\end{table*}


\begin{table*}[!htbp]
    \centering
\hspace{-1.2cm}
\small
\setlength{\tabcolsep}{1.7pt}
    \begin{tabular}{|l|r|r|r|r|r|r|r|}
        \hline
        \textbf{} & \textbf{2018} & \textbf{2019} & \textbf{2020} & \textbf{2021} & \textbf{2022} & \textbf{Thereafter} & \textbf{Total} \\
        \hline
        Property mortgages and other loans & 153593 & 42289 & 703018 & 11656 & 208003 & 1656623 & 2775182 \\
        MRA facilities & 90809 & 0 & 0 & 0 & 0 & 0 & 90809 \\
        Revolving credit facility & 0 & 0 & 0 & 0 & 0 & 40000 & 40000 \\
        Unsecured term loans & 0 & 0 & 0 & 0 & 0 & 1500000 & 1500000 \\
        Senior unsecured notes & 250000 & 0 & 250000 & 0 & 800000 & 100000 & 1400000 \\
        Trust preferred securities & 0 & 0 & 0 & 0 & 0 & 100000 & 100000 \\
        Capital lease & 2387 & 2411 & 2620 & 2794 & 2794 & 819894 & 832900 \\
        Ground leases & 31049 & 31066 & 31436 & 31628 & 29472 & 703254 & 857905 \\
        Estimated interest expense & 226815 & 218019 & 184376 & 163648 & 155398 & 281694 & 1229950 \\
        Joint venture debt & 200250 & 717682 & 473809 & 449740 & 223330 & 2119481 & 4184292 \\
        Total & 954903 & 1011467 & 1645259 & 659466 & 1418997 & 7320946 & 13011038 \\
        \hline
    \end{tabular}
    \caption{Loans and Liabilities, Loans, MultiHiertt example Table 1}
    \label{tab:loans_tab1_multi}
\end{table*}

\begin{table}[H]
\small
    \centering
    \begin{tabular}{|l|r|}
        \hline
        \textbf{Year} & \textbf{Amount (\$)} \\
        \hline
        2007 & 56499000 \\
        2008 & 46899000 \\
        2009 & 39904000 \\
        2010 & 33329000 \\
        2011 & 25666000 \\
        Later Years & 128981000 \\
        \hline
    \end{tabular}
    \caption{Aggregate Minimum Lease Payments, Lease Payments FinQA}
    \label{tab:lease_payments_finqa}
\end{table}

\bigskip
\noindent\textbf{Total Debt Overview: for FinQA example.}  Total debt at July 1, 2006 was \$1,762,692,000, of which approximately 75 was at fixed rates averaging 6.0 with an average life of 19 years, and the remainder was at floating rates averaging 5.2. Certain loan agreements contain typical debt covenants to protect noteholders, including provisions to maintain the company’s long-term debt to total capital ratio below a specified level. Sysco was in compliance with all debt covenants at July 1, 2006. 

The fair value of Sysco’s total long-term debt is estimated based on the quoted market prices for the same or similar issues or on the current rates offered to the company for debt of the same remaining maturities. The fair value of total long-term debt approximated \$1,669,999,000 at July 1, 2006 and \$1,442,721,000 at July 2, 2005, respectively. As of July 1, 2006 and July 2, 2005, letters of credit outstanding were \$60,000,000 and \$76,817,000, respectively.

\bigskip
\noindent\textbf{Leases:  for FinQA example.} Although Sysco normally purchases assets, it has obligations under capital and operating leases for certain distribution facilities, vehicles, and computers. Total rental expense under operating leases was \$100,690,000, \$92,710,000, and \$86,842,000 in fiscal 2006, 2005, and 2004, respectively. Contingent rentals, subleases, and assets and obligations under capital leases are not significant. Aggregate minimum lease payments by fiscal year under existing non-capitalized long-term leases are as follows:

\section{{Detailed Error Analysis}}
\label{sec:error-analysis-details}

{We conduct a detailed error analysis across six datasets to identify the primary failure modes in pipeline‑based table QA. As shown in plot \ref{fig:error-dist} evidence extraction errors dominate in most datasets, often occurring before reasoning or code execution can contribute. However, we also observe notable secondary errors particularly in reasoning (e.g., TAT‑QA) and code (e.g., WikiTQ) which vary by dataset structure and modality.}

{\subsection*{Dataset‑specific Observations}}

{\paragraph{HybridQA.}
Most errors result from incorrect row/column selection, driven by surface‑level matches to look‑alike strings (e.g., “season,” “division”) and missed disambiguators (e.g., years or suffixes like “(q)”). These tokens frequently appear in adjacent cells or parentheses, making shallow matches more likely. A small number of reasoning errors stem from failure to disambiguate linked entities across tables. Code issues are rare due to minimal programmatic computation.}

{\paragraph{HiTABs.}
Models often fail due to header ambiguity and dense, repetitive tabular layouts. Errors stem from misidentified rows/columns and unaccounted qualifiers like ranges or footnotes. Since the task is primarily table lookup, downstream reasoning errors are minimal, and code is not a significant factor.}

{\paragraph{MultiHiertt (Multi).}
High error rate in evidence selection is attributed to multi‑hop grounding and similar‑looking headers across tables. Subtle distinctions in qualifiers or column semantics are frequently overlooked. The remaining errors are due to misinterpretation of multi‑hop logic, where the model fails to chain intermediate inferences.}

{\paragraph{TAT‑QA.}
While evidence errors are common, reasoning mistakes form a large minority (36\%), often caused by temporal mismatches (e.g., Q1 vs.\ FY) or incorrect unit normalization (e.g., billions vs.\ millions). Models struggle to align period‑based values or compute correct numerical operations even with correct evidence.}

{\paragraph{FETAQA.}
Evidence‑level failures persist due to repeated phrases across seasons, clubs, and divisions. Parenthetical markers (e.g., “(q)”) in headers lead to grounding mismatches. Some errors stem from reasoning failures, particularly aggregation mismatches or improper scoping across semi‑structured tables. A few errors involve minor code missteps, such as summing incorrect subsets.}


{\paragraph{WikiTQ.}
Unlike others, code generation errors dominate (44\%). The model often produces incorrect filters or aggregation logic due to brittle parsing of semi‑structured HTML‑derived tables. Even when correct evidence is identified, the final output is wrong due to mis‑joins, faulty parsing of footnotes, or wrong aggregation. Evidence errors (40\%) and reasoning mistakes (16\%) persist but are less frequent.}

\medskip
{The analysis reveals that evidence selection remains the primary bottleneck across most datasets. However, reasoning errors are increasingly relevant in multi‑hop or temporal computation tasks (TAT‑QA), and code execution errors emerge as a major challenge in semi‑structured, programmatic tasks like WikiTQ. Addressing early‑stage grounding \emph{and} late‑stage execution together is critical for end‑to‑end accuracy.}

\section{DataSet Overview}
\label{dat: overview of dataset}
\begin{enumerate}
\setlength\itemsep{0em}
\item\textbf{FeTaQA\cite{nan2021fetaqafreeformtablequestion} :} A Wikipedia-based table QA dataset that requires generating long-form answers by integrating multiple discontinuous facts and reasoning across structured tables. \textbf{Temporal Questions: 1,582}
\item\textbf{FinQA\cite{chen-etal-2021-finqa} :} A financial QA dataset from reports, requiring expert-verified multi-step numerical reasoning and gold reasoning programs for explainability. \textbf{Temporal Questions: 962}
\item\textbf{HiTab\cite{cheng-etal-2022-hitab} :} A cross-domain QA and NLG dataset featuring hierarchical tables, analyst-authored questions, and fine-grained annotations for complex numerical reasoning. \textbf{Temporal Questions: 897}
\item\textbf{HybridQA\cite{chen-etal-2020-hybridqa} :} A QA dataset requiring reasoning over Wikipedia tables and linked free-form text, demanding both tabular and textual data for accurate answers. \textbf{Temporal Questions: 1,528}
\item\textbf{MultiHierTT\cite{zhao2022multihierttnumericalreasoningmulti} :} A financial QA benchmark requiring reasoning over multiple hierarchical tables and long unstructured text, with detailed multi-step numerical reasoning annotations. \textbf{Temporal Questions: 1,587}
\item\textbf{Squall\cite{shi-etal-2020-potential} :} An extension of WikiTableQuestions with manually created SQL equivalents and fine-grained alignments, supporting structured query reasoning in tabular environments. \textbf{Temporal Questions: 774}
\item\textbf{TAT-QA\cite{zhu2021tatqaquestionansweringbenchmark} :} A financial QA dataset requiring reasoning over both tabular and textual data, involving operations like arithmetic, counting, and sorting for quantitative and qualitative analysis. \textbf{Temporal Questions: 2,244}
\item\textbf{WikiTableQ\cite{pasupat2015compositionalsemanticparsingsemistructured} :} A Wikipedia-based QA dataset with trivia-style questions requiring factual and numerical reasoning over tables with at least 8 rows and 5 columns. \textbf{Temporal Questions: 1,504}
\end{enumerate}


\end{document}